\documentclass{article}

\usepackage{PRIMEarxiv}

\usepackage[utf8]{inputenc} 
\usepackage[T1]{fontenc}    
\usepackage{hyperref}       
\usepackage{url}            
\usepackage{booktabs}       
\usepackage{amsfonts}       
\usepackage{nicefrac}       
\usepackage{microtype}      
\usepackage{lipsum}
\usepackage{fancyhdr}       
\usepackage{graphicx}       
\graphicspath{{media/}}     
\usepackage{amsmath}
\usepackage{markdown}
\usepackage[ruled,vlined]{algorithm2e}
\usepackage[table]{xcolor}
\SetKwComment{Comment}{$\triangleright$\ }{}
\usepackage{subcaption}
\usepackage{soul}
\usepackage{arydshln}

\title{Learning from Few Examples: \linebreak A Summary of Approaches to Few-Shot Learning
}

\author{
  Archit Parnami and Minwoo Lee \\
  Department of Computer Science \\
  The University of North Carolina at Charlotte \\
  Charlotte, NC, USA\\
  \texttt{\{aparnami, minwoo.lee\}@uncc.edu} \\
}

\begin{document}
\maketitle

\begin{abstract}
Few-Shot Learning refers to the problem of learning the underlying pattern in the data just from a few training samples. Requiring a large number of data samples, many deep learning solutions suffer from data hunger and extensively high computation time and resources. Furthermore, data is often not available due to not only the nature of the problem or privacy concerns but also the cost of data preparation. Data collection, preprocessing, and labeling are strenuous human tasks. Therefore, few-shot learning that could drastically reduce the turnaround time of building machine learning applications emerges as a low-cost solution. This survey paper comprises a representative list of recently\footnote{Until January 2020.} proposed few-shot learning algorithms. Given the learning dynamics and characteristics, the approaches to few-shot learning problems are discussed in the perspectives of meta-learning, transfer learning, and hybrid approaches (i.e., different variations of the few-shot learning problem).
\end{abstract}


\vfill
\pagebreak
\setcounter{tocdepth}{4}
\setcounter{secnumdepth}{5}
\tableofcontents
\listoftables
\pagebreak

\section{Introduction}
The field of Artificial Intelligence (AI) has been through ups and downs since its inception in the 1950s. Yet, the last few years have been marked by exceptional progress in the field of AI.  Much of this progress can be attributed to recent advances in “deep learning” characterized by learning large neural network-style models with multiple layers of representation.  These models have shown great performance in a variety of tasks with large amounts of labeled data in image classification \cite{he2016deep}, machine translation \cite{GoogleNMT} and speech modeling \cite{oord2016wavenet}. However, these achievements have relied on the fact that the optimization of these deep, high-capacity models requires many iterative updates across many labeled examples. This type of optimization breaks down in the small data regime where we want to learn from very few labeled examples.  In contrast, humans can quickly learn to solve a new problem just from a few examples. For instance, given a few photos of a stranger, one can easily identify the same person from a large number of photos.  This is not only due to the human mind's computational power but also to its ability to synthesize and learn new information from previously learned information. For example, if a person has a skill for riding a bicycle, that skill can prove helpful when learning to ride a motorcycle.

\begin{figure*}[ht]
    \centering
    \includegraphics[keepaspectratio,width=0.8\textwidth]{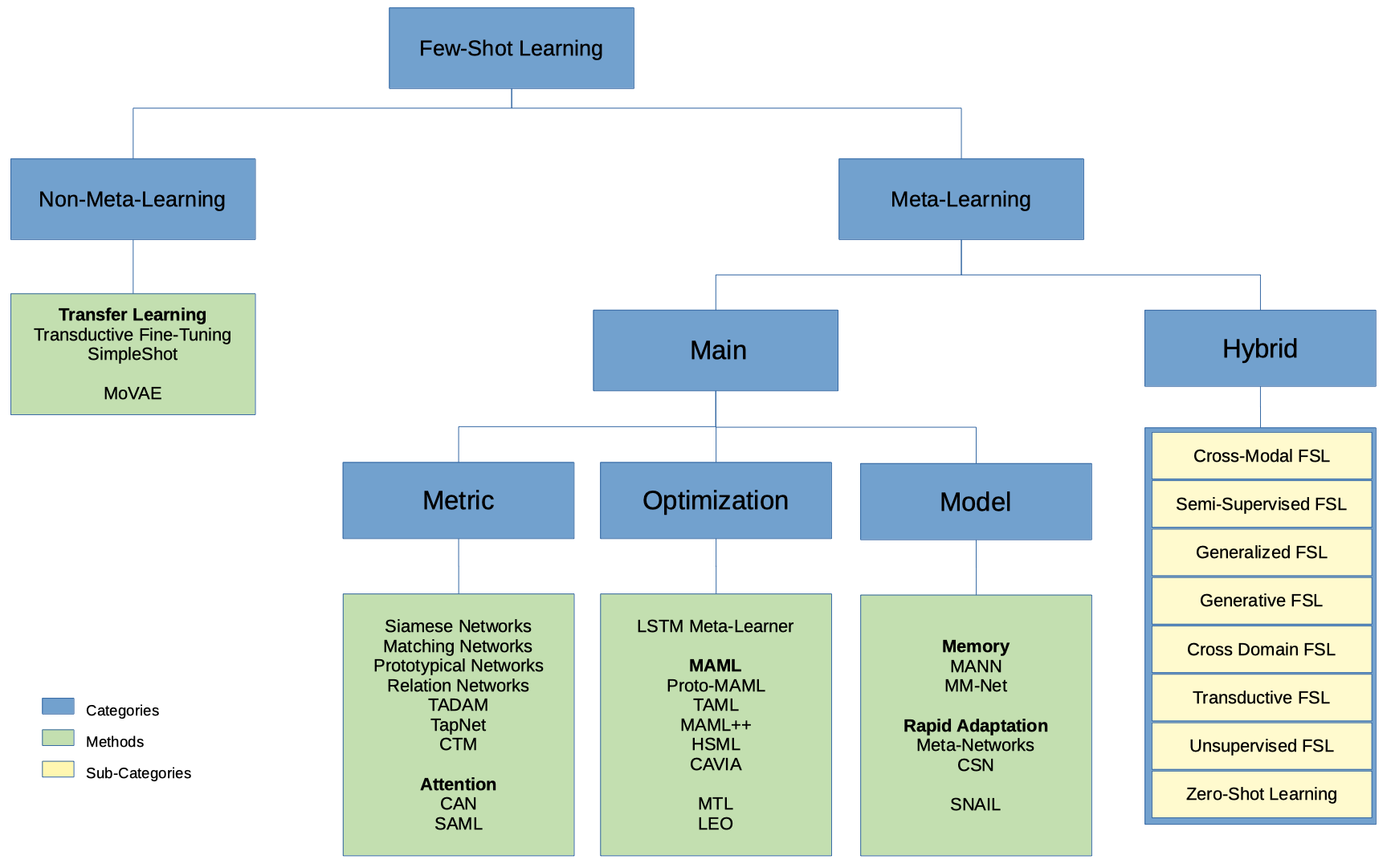}
    \caption{Approaches to FSL are categorized into Meta-Learning-based FSL and Non-Meta-Learning-based FSL. The three main meta-learning approaches are: metric-based, optimization-based and model-based meta-learning. Furthermore, variations of the FSL problem which use meta-learning are categorized as hybrid approaches.}
    \label{fig:classification}
\end{figure*}

Recent years have seen a rise of research attempting to bridge this gap between human-like learning and machine learning, which has given birth to this new sub-field of Machine Learning known as \textbf{Few-Shot Learning (FSL)} i.e., \textbf{the ability of machine learning models to generalize  from few training examples}.  When there is only one example to learn from, FSL is also referred to as One-Shot Learning. The motivation for FSL lies in the fact that models excelling at this task would have many useful applications. First, we would not be required to collect thousands of labeled examples to attain a reasonable performance on a new task which would help alleviate the data-gathering effort and reduce the computation costs  and time spent in training a model. Furthermore, in many fields, data is hard or impossible to acquire due to reasons such as privacy and safety.  Models that generalize from a few examples would be able to capture this type of data effectively. Therefore in this survey, we study various approaches that have been recently proposed in an attempt to solve the problem of Few-Shot Learning. We categorize the FSL approaches as seen in Figure \ref{fig:classification}.  

Much of the work in this survey is inspired from prior works that attempted to summarize the field of few-shot learning.  Wang et al. \cite{Wang2019GeneralizingFA} defines the core issue of FSL as an unreliable empirical risk minimizer that makes it a hard problem.  Chen et al. \cite{Chen2019ACL} present a comparative analysis of several representative few-shot classification algorithms. Similar to ours,  Weng \cite{lilianweng} discusses meta-learning approaches to FSL. In addition, we discuss non-meta-learning and hybrid meta-learning approaches to FSL. We also extend the discussion of main meta-learning approaches by including the recent state-of-the-art methods.

The rest of the survey is organized as follows. Section \ref{sec:background} covers meta-learning which is a prerequisite to understand recent FSL methods. Section \ref{sec:fsl} defines the few-shot classification problem, introduces the available datasets, and divides the approaches to FSL into two sub-sections: Meta-Learning-based FSL (Section \ref{sec:meta-fsl} and Non-Meta-Learning-based FSL (Section \ref{sec:non-meta-fsl}). Next, we list the performance of various FSL methods and progress made so far in Section \ref{sec:progress}. Finally, we conclude with a discussion on challenges involved in Section \ref{sec:discussion}. 

\section{Background} \label{sec:background}
In this section, we discuss the necessary background required for understanding the recent few-shot learning algorithms.

\subsection{Meta-Learning}
Meta-Learning \cite{schmidhuber:1987:srl,Schaul:2010} or Learning to Learn \cite{Thrun&Pratt} has been the basic technique which most few-shot learning algorithms employ. Motivated by human development theory, meta-learning, a sub-field of machine learning, focuses on learning priors from previous experiences that can lead to efficient downstream learning of new tasks. For instance, a simple learner learns a single classification task, but a meta-learner gains an understanding of learning to solve a classification task by exposing itself to multiple similar classification tasks. Hence when presented with a similar but new task, the meta-learner could solve it quickly and better than a simple learner which has no prior experience in solving the task. A meta-learning procedure generally involves learning at two levels, within and across tasks. First, rapid learning occurs \textit{within} a task, for example, learning to accurately classify within a particular dataset. Next, this learning is guided by knowledge accrued more gradually \textit{across} tasks, which captures the way task structure varies across target domains. 

Meta-Learning can be different from similar approaches such as transfer learning, multi-task learning, or ensemble learning. In Transfer Learning \cite{pan2009survey}, a model is trained on a single task known as the source task in the source domain where the sufficient training data is available. This trained model is then again retrained or finetuned on another single task known as the target task in the target domain. The transfer of knowledge occurs from the source task to the target task. Thus more similar the two domains are the better it performs . Multi-Task Learning \cite{Caruana1998}, involves learning multiple tasks simultaneously.  It starts from no prior experience and attempts to optimize over solving multiple tasks at the same time.  On the other hand, Ensemble Learning \cite{Polikar:2009} is the process by which multiple models, such as classifiers or experts, are strategically generated and combined to solve a particular task. In contrast, meta-learner first gathers experience across multiple similar tasks and then use that experience to solve new tasks. Nonetheless, these techniques can be and often are meaningfully combined with meta-learning systems. We provide the formal definition for the meta-learning problem and explain it with the help of an example.

\subsubsection{Problem Definition} \label{sec:problem-definition}

In a typical supervised learning setting, we are interested in a task $\mathcal{T}$  with a dataset $\mathcal{D} = \{(x_k,y_k)\}_{k=1}^n$ with $n$ data samples. We usually split $\mathcal{D}$ into $\mathcal{D}^{train}$ and $\mathcal{D}^{test}$ such that:

\begin{equation} \label{eq:dtrain}
  \mathcal{D}^{train} = \{(x_k,y_k)\}_{k=1}^t   
\end{equation}

and

\begin{equation} \label{eq:dtest}
  \mathcal{D}^{test} = \{(x_k,y_k)\}_{k=t+1}^n,
\end{equation}

where $t$ denotes the number of training samples. We optimize parameters $\theta$ on the training set $\mathcal{D}^{train}$ and evaluate its generalization performance on the test set $\mathcal{D}^{test}$. Thus the learning problem here is to approximate the function $f$ with parameters $\theta$ as 
\footnote{We omit sample subscript $k$ for simplicity in the following discussion}:

\begin{equation} \label{eq:simple-predict}
    y \approx f(x; \theta) \text{ where }   (x,y) \in \mathcal{D}^{test}
\end{equation}

and

\begin{equation} \label{eq:simple-theta}
  \theta = \text{arg} \min\limits_{\theta} \sum\limits_{(x,y) \in \mathcal{D}^{train}} \mathcal{L}(f(x, \theta), y)  
\end{equation}

where $\mathcal{L}$ is any loss function measuring the error between the prediction $f(x,\theta)$ and the true label $y$. 

In meta-learning, we have a distribution  $p(\mathcal{T})$ of task $\mathcal{T}$.  A  meta-learner learns from a set of training tasks $\mathcal{T}_{i} \stackrel{train}{\sim} p(\mathcal{T})$ and is evaluated on a set of testing tasks  $\mathcal{T}_{i} \stackrel{test}{\sim} p(\mathcal{T})$.  Each of these task has its own dataset $\mathcal{D}_{i}$ where $\mathcal{D}_i = \{\mathcal{D}^{train}_i, \mathcal{D}^{test}_i\}$.

Let us denote the set of training tasks as $\mathcal{T}_{meta-train} = \{\mathcal{T}_1, \mathcal{T}_2, ..... , \mathcal{T}_n\}$ and the set of testing tasks as $\mathcal{T}_{meta-test} = \{\mathcal{T}_{n+1}, \mathcal{T}_{n+2}, ..... , \mathcal{T}_{n+k}\}$. Correspondingly, the training dataset for the meta-learner will be $\mathcal{D}_{meta-train} = \{\mathcal{D}_1, \mathcal{D}_2, ..... , \mathcal{D}_n\}$ and the testing dataset will be $\mathcal{D}_{meta-test} = \{\mathcal{D}_{n+1}, \mathcal{D}_{n+2}, ..... , \mathcal{D}_{n+k}\}$. 

The parameters $\theta$ of the meta-learner are optimized on $D_{meta-train}$ and its generalization performance is tested on $D_{meta-test}$.  Then, the meta-learning problem is to approximate the function $f$ with parameters $\theta$ as:

\begin{equation} \label{eq:meta-predict}
    y  \approx f(\mathcal{D}^{train}_{i}, x; \theta) \text{ where } (x,y) \in \mathcal{D}^{test}_{i} 
\end{equation}

and

$$
    \mathcal{D}_i = \{\mathcal{D}^{train}_{i}, \mathcal{D}^{test}_i\} \text{ where } \mathcal{D}_{i} \in \mathcal{D}_{meta-test}
$$

i.e., $\mathcal{D}_i$ is the dataset for a test task $\mathcal{T}_{i}$ sampled from $\mathcal{T}_{meta-test}$.  Then the optimal model 
parameters are obtained as:

\begin{equation} \label{eq:meta-theta}
    \theta = \text{arg} \min\limits_{\theta} \sum\limits_{\mathcal{D}_i \in \mathcal{D}_{meta-train}}  \sum\limits_{(x,y) \in \mathcal{D}^{test}_i} \mathcal{L}(f(\mathcal{D}^{train}_i,x; \theta), y).
\end{equation}

\begin{figure*}[ht]
    \centering
    \includegraphics[keepaspectratio,width=\textwidth]{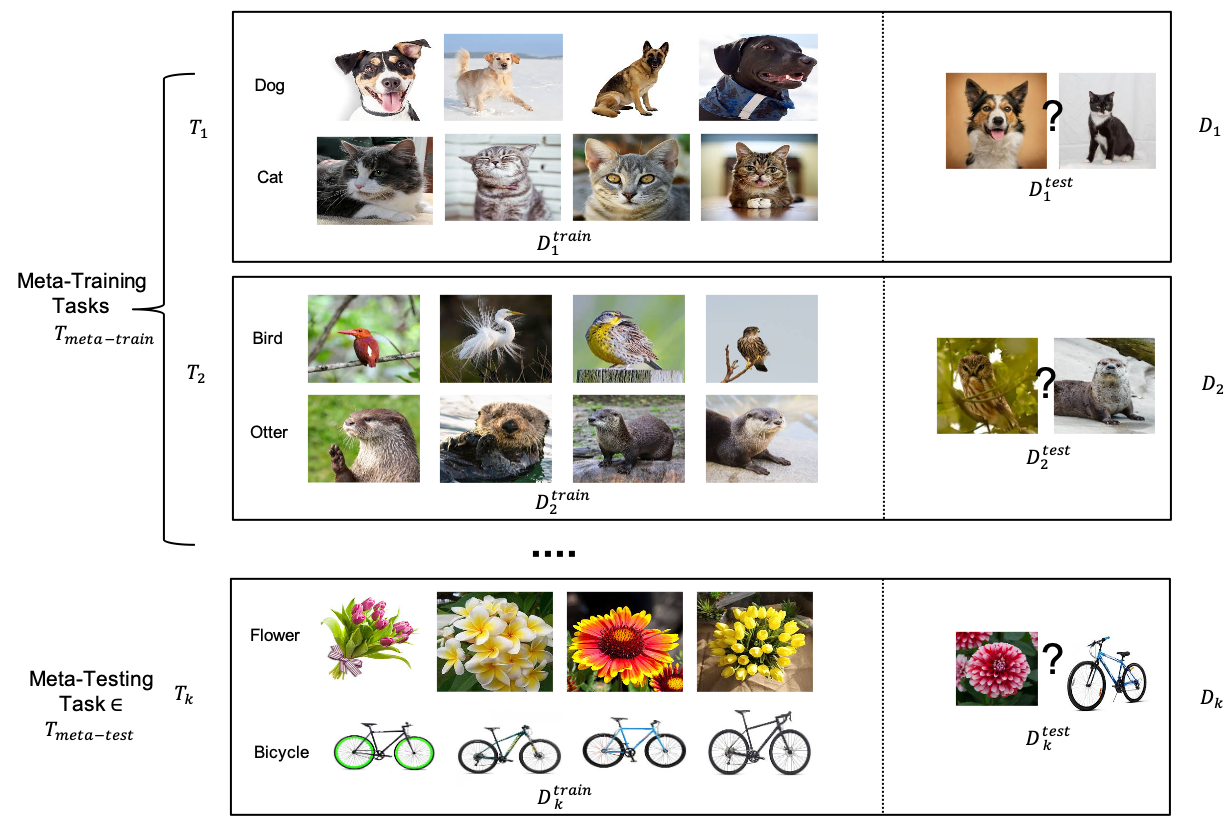}
    \caption{\textbf{Meta-Learning Example Setup}. Each task $\mathcal{T}_i$ is a binary classification task with a training set $\mathcal{D}^{train}_i$ and test set $\mathcal{D}^{test}_i$. During meta-training, the labels for samples in $\mathcal{D}^{test}_i$ is known and the goal of meta-learner is to find optimal $\theta$ as per equation \ref{eq:meta-theta}. During meta-testing, new task with unseen categories is presented and the labels are predicted as per equation \ref{eq:meta-predict} . }
    \label{fig:meta-setup}
\end{figure*}

That is the meta-learner learns parameters $\theta$ such that given a task $\mathcal{T}_i \sim p(\mathcal{T})$, its performance on its test data $\mathcal{D}^{test}_i$ given its training data $\mathcal{D}^{train}_i$ would be optimal. Figure \ref{fig:meta-setup} demonstrates a setup for example meta-learning problem. 

\subsubsection{Nomenclature}
In meta-learning and few-shot learning literature, certain notations and terms are used interchangeably. Table \ref{table:nom} lists these terms and their equivalent usage. Notation \textbf{A} is more commonly used in optimization-based meta-learning literature (Section \ref{sec:optim}) while notation \textbf{B} is used when discussing metric-based meta-learning methods (Section \ref{sec:metric}). Additionally, Table \ref{table:symbols} lists the commonly used symbols in this survey.

\begin{table}[!ht]
    \centering
    \begin{tabular}{llll}
    \hline
        \textbf{Notation A} & \textbf{Term A} & \textbf{Notation B} & \textbf{Term B} \\ \hline
        $\mathcal{D}^{train}_i$     & Training set for task $\mathcal{T}_i$ & $S_{i}$        & Support Set for task $\mathcal{T}_i$ \\
        $\mathcal{D}^{test}_i$     & Test set for task $\mathcal{T}_i$     & $Q_{i}$        & Query Set for task $\mathcal{T}_i$   \\
        $\mathcal{D}_{meta-train}$    & Meta-training set           & $\mathcal{D}_{train}$    & Training Set                  \\
        $\mathcal{D}_{meta-test}$     & Meta-testing set            & $\mathcal{D}_{test}$     & Test Set                      \\
    \hline
    \end{tabular}
    \caption{Nomenclature}
    \label{table:nom}
\end{table}
\begin{table}[!ht]
    \centering
    \begin{tabular}{llp{2cm}}
        \hline
        \textbf{Symbol} & \textbf{Meaning}  & \textbf{Context} \\ \hline
        $\mathcal{T}_{i}$  & Task $i$ &  \\
        $\mathcal{L}$ & Loss function &  \\
        $(x_k,y_k)$ & Input-Output pair &  \\
        $f_{\theta}$ & Model (function) with parameters $\theta$ & \\
        $g_{\theta_1}$ & Embedding function & Sec. \ref{sec:metric} \\
        $d_{\theta_2}$ or $d$ & Distance function  & Sec. \ref{sec:metric} \\
        $g_{\phi}$ & Meta-Learning model with parameters $\phi$  & Sec. \ref{sec:optim} \\
        $P_{\theta}(y|x)$ &  Output probability of $y$ for input $x$ using model parameters $\theta$ \\
        $k_{\theta}(x_1,x_2)$ & Kernel function measuring similarity between two vectors $x_1$ and $x_2$ & Table \ref{table:meta-learning-methods} \\
        $\sigma$ & Softmax function  \\
        $\alpha, \beta$ & Learning rates  \\
        $w$ & Weights & \\
        $\mathbf{v}_c$ & Prototype of class $c$ & Eq. \ref{eq:prototype} \\
        $C$ & Set of classes present in  S \\
        $S^c$ & Subset of S containing all elements $(x_k,y_k)$ such that $y_k = c$  \\
        $\oplus$ & Concatenation operator & Table \ref{table:metric-ml} \\
        $B$ & Number of batches $(X_b,Y_b)$ sampled in inner-loop for a randomly sampled task $\mathcal{T}_i$ & Table \ref{table:optim} \\
        $I$ & Number of tasks $\mathcal{T}_i$ sampled in inner-loop & Table \ref{table:optim} \\
        $J$ & Number of outer-loop iterations & Table \ref{table:optim} \\
        \hline
    \end{tabular}
    \caption{Symbols}
    \label{table:symbols}
\end{table}

\section{Few-Shot Learning} \label{sec:fsl}
Much of the recent progress in FSL has come through meta-learning. Therefore we first divide the approaches to FSL into two categories: meta-learning-based FSL and non-meta-learning-based FSL.  Also, most of these approaches that we are about to discuss were developed with the perspective of solving the few-shot image classification problem. However, they are still applicable for solving other problems such as regression, object detection, segmentation, online recommendation, reinforcement learning, etc. We discuss the approaches to the few-shot image classification problem in this section and the applications to other domains in Section \ref{sec:progress} .

\subsection{The Few-Shot Classification Problem} \label{sec:fsc}
Consider the task $\mathcal{T}$ (defined in Section \ref{sec:problem-definition}) as a classification task where $x$ is input and $y$ is the output label.  The objective is to approximate the function $f$ (Eqn. \ref{eq:simple-predict}) with parameters $\theta$ (Eqn. \ref{eq:simple-theta}) . This is generally possible when we have sufficient training data $\mathcal{D}^{train}$ (Eqn. \ref{eq:dtrain}) i.e., $t$ is a large number. However, when $t$ is small, it becomes difficult to approximate the function $f$ so that it has good generalization performance over $\mathcal{D}^{test}$ (Eqn. \ref{eq:dtest}). This can be referred to as a \textit{few-shot classification problem} as the number of examples (shots) are too few to learn a good model. 

Usually people define few-shot classification task as a standard \textbf{M-way-K-shot} task \cite{Matching2016,MAML2017}, where M is the number of classes and K is the number of examples per class present in $\mathcal{D}^{train}$.  Usually K is a small number (ex., 1,5,10) and  $|\mathcal{D}^{train}| = M \times K$. The performance is measured by a loss function $\mathcal{L}(\hat{y}, y)$ defined over the prediction $\hat{y} = f(x, \theta)$ and the ground truth $y$. 

The \textbf{M-way-K-shot} tasks are usually sampled from a larger dataset with classes much higher in number than \textit{M}. Table \ref{table:datasets} lists such commonly used datasets for conducting experiments for few-shot classification. 

\begin{table}[!ht]
    \centering
    \begin{tabular}{lllp{6.7cm}}
    \hline
        \textbf{Dataset} & \textbf{Number of classes} & \textbf{Samples per class} & \textbf{Description} \\ \hline
        Omniglot \cite{Lake1332} & 1623 & 20 & Handwritten characters from different languages. \\
        \textit{mini}ImageNet \cite{Matching2016} & 100 & 600 & 100 classes randomly sampled from ImageNet. \\
        FC100 \cite{Oreshkin2018TADAMTD} & 100 & 600 & Derived from CIFAR100 for FSL. \\
        \textit{tiered}ImageNet \cite{Ren2018MetaLearningFS} & 608 & 1280 (avg.) & Like \textit{mini}ImageNet but ensures that there is a wider degree of separation between training, validation and test classes. \\
    \hline
    \end{tabular}
    \caption{Common datasets used for Few-Shot Learning}
    \label{table:datasets}
\end{table}

\subsection{Meta-Learning-based Few-Shot Learning} \label{sec:meta-fsl}
The objective of meta-learning is to approximate the function $f$ with parameters $\theta$ such that the performance on any task $\mathcal{T}_i$ randomly sampled from the task distribution $p(\mathcal{T})$ is optimal (equation \ref{eq:meta-predict}). We use this strategy for FSL such that the distribution $p(\mathcal{T})$ is now a distribution of few-shot tasks and each task $\mathcal{T}_i$ is a few-shot task. For example, consider the M-way-K-shot few-shot classification (FSC) task. During training, we meta-learn a prior $\theta$ over a distribution of M-way-K-shot FSC tasks so that at test time we can solve for a new M-way-K-shot FSC task. 

Meta-Learning-based FSL can be classified into three approaches \cite{vinyals}: metric-based, optimization-based and model-based.  Further, various meta-learning-based hybrid approaches were proposed to handle FSL problems such as cross-domain FSL, generalized FSL, etc. We discuss the three main approaches in Section \ref{sec:main} and hybrid approaches in Section \ref{sec:hybrid}.

\subsubsection{Main Approaches} \label{sec:main}
Consider a task $\mathcal{T}$ with support set $S$ and query set $Q$. Let $f$ be a few-shot classification model with parameters $\theta$. Then for  $(x,y) \in Q$, the meta-learning approaches to FSL can be differentiated in the way they model the output probability $P_{\theta}(y|\mathbf{x})$ \cite{vinyals} (Table \ref{table:meta-learning-methods}). 

\begin{table}[!ht]
    \centering
    \begin{tabular}{p{2cm}p{4cm}p{4cm}p{4cm}}
        \hline
         & \textbf{Metric-based} & \textbf{Optimization-based}  & \textbf{Model-based} \\ \hline
         \textbf{Key idea} & Metric Learning \cite{MetricLearning} & Gradient Descent  & Memory; RNN \\
         \\
         \textbf{How $P_{\theta}(y|x)$ is modeled?} & $$\sum_{(x_k,y_k) \in S} k_{\theta}(x,x_k) y_k,$$ &  $$P_{\theta^{'}}(y|x),$$ where  $\theta^{'}$ =  $g_{\phi}(\theta, S)$ & $$f_{\theta}(x,S).$$  \\
         \\
        \textbf{Advantages} & Faster Inference. &  Offers flexibility to optimize in dynamic environments.  &  Faster inference with memory models. \\
                            & Easy to deploy. &  $S$ can be discarded post-optimization. &  Eliminates the need for defining a metric or optimizing at test. \\
        \\
        \textbf{Disadvantages} &  Less adaptive to optimization in dynamic environments. &  Optimization at inference is undesirable for real-world deployment. &  Less efficient to hold data in memory as $S$ grows. \\
                                & Computational complexity grows linearly with size of $S$ at test. &  Prone to overfitting. &  Hard to design. \\
        \hline
    \end{tabular}
    \caption{Meta-Learning Approaches}
    \label{table:meta-learning-methods}
\end{table}

\pagebreak

\paragraph{Metric-based Meta-Learning} \label{sec:metric} \mbox{}\\
Metric Learning \cite{MetricLearning} is the task of learning a distance function over data samples. Consider two image-label pair $(x_1,y_1)$ and $(x_2,y_2)$ and a distance function $d$ to measure the distance between two images. If we were to assign a label to a query image $x_3$, we could compute the two distances $d(x_1, x_3)$ and $d(x_2, x_3))$ and assign the label corresponding to the image with a shorter distance, which is also the key idea in nearest neighbors algorithms (k-NN). However, with high dimensional inputs such as images, we typically use an embedding function $g$ to transform the input to a lower dimension before computing the distances:

$$g \colon  \mathbb{R}^n \to \mathbb{R}^m \text{  where } n > m.$$ 

\begin{figure*}[ht]
    \centering
    \includegraphics[keepaspectratio,width=\textwidth]{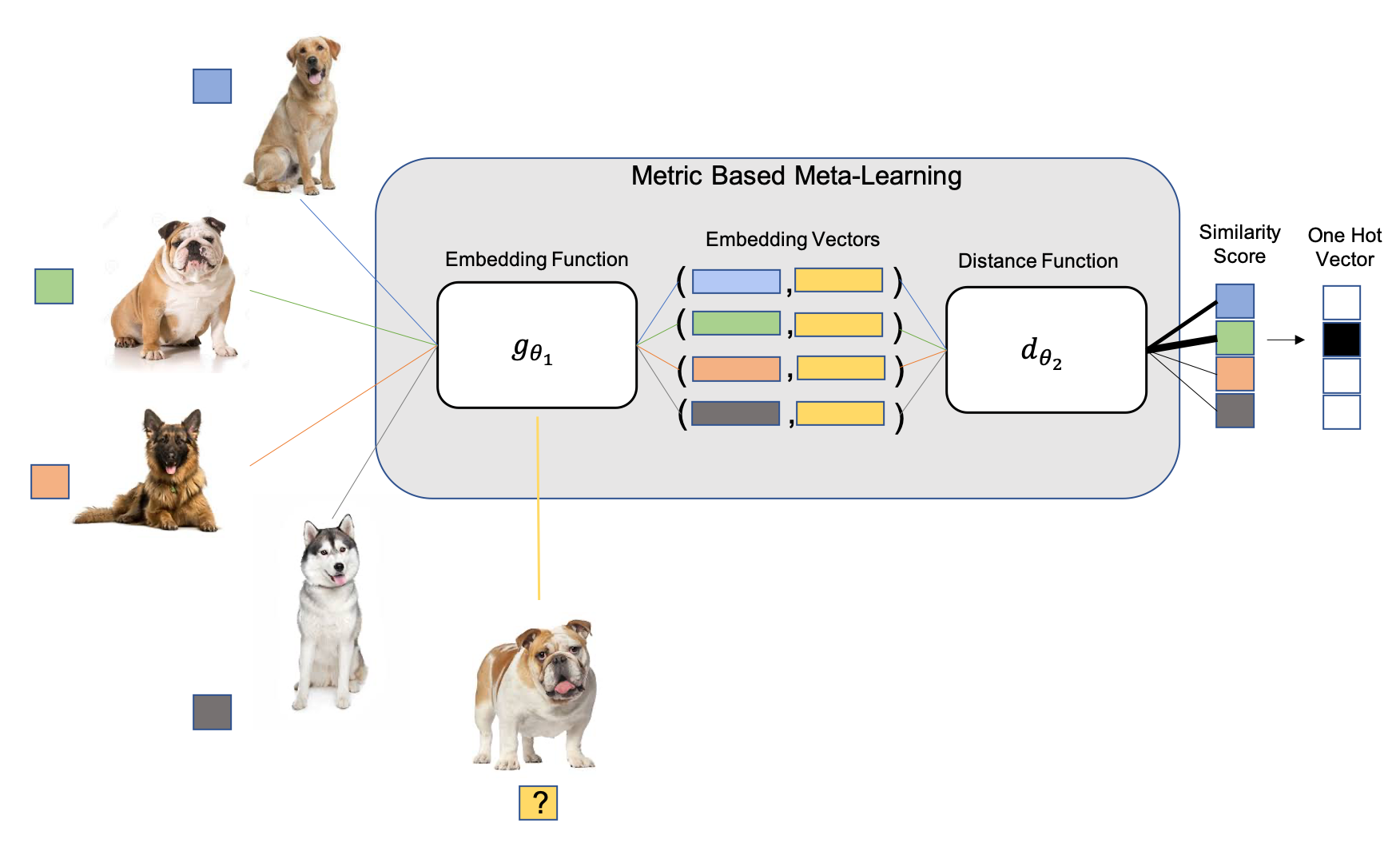}
    \caption{Example metric-based meta-learning setup for a 4-way-1-shot classification task. The embedding function $g_{\theta_1}$ outputs the embedding vectors for support images (labeled) and the query image (unlabeled, denoted by '?'). Distance function $d_{\theta_2}$ measures the distance between support and query vectors to output a similarity score. }
    \label{fig:metric}
\end{figure*}

Therefore, the core idea behind metric-based few-shot learning is to leverage meta-learning architecture to either learn an embedding function $g(;\theta_1)$ (parameterized by a neural network with parameters $\theta_1$) given a distance function $d$ (such as euclidean distance) or to learn both the embedding function $g(;\theta_1)$ (with parameters $\theta_1$) and the distance function $d(;\theta_2)$ (usually parameterized by another neural network with parameters $\theta_2$). This is illustrated in Figure \ref{fig:metric} .

Training proceeds by randomly sampling M-way-K-shot episodes from the training set. Each episode has a support set and a query set. The average error computed on query sets across multiple training few-shot episodes is used to update the parameters of the embedding function and the distance function (if any). Finally,  new M-way-K-shot episodes are sampled from the testing set to evaluate the performance of the network. This episodic training paradigm is explained in Algorithm \ref{alg:episodic} .

\begin{algorithm}

\DontPrintSemicolon
\SetAlgoLined

\textbf{Given:} In dataset $D$, $n$ is the number of examples, $N$ is the set of classes, $N_{train}$ is the set of classes used for training, $N_{test}$ is the set of classes used for testing, $M < N_{train}$ is the number of classes per episode, $K$ is the number of support examples per class, $Q$ is the number of query examples per class. $RandomSample(A,B)$ denotes a set of $B$ elements chosen uniformly at random from set $A$,  $|N_{train}| + |N_{test}| = |N|$ and $N_{train} \cap N_{test} = \emptyset$.
\BlankLine

\textbf{Input:} $D = \{(x_1, y_1), ...,  (x_n, y_n)\}$  where $y_i \in \{1,...,N\}$. $D^c$ denotes the subset of D containing all elements $(x_i, y_i)$ such that $y_i = c$.
\BlankLine

\textbf{Training}: \Comment*[r]{M-way K-shot training episodes}
\BlankLine

\While {True}
{   
    \Comment*[r]{1. Constructing Task}
    \BlankLine
    $C \leftarrow RandomSample(N_{train}, M)$   \Comment*[r]{Sample M classes}  
    $S \leftarrow \{\}$ \Comment*[r]{Support set}
    $Q \leftarrow \{\}$ \Comment*[r]{Query set}
    \For{$c$ in $C$}  
    {
        $S^c \leftarrow RandomSample(D^c, K)$  \Comment*[r]{Sample K support}   
        $Q^c \leftarrow RandomSample(D^c \setminus  S^c, Q)$  \Comment*[r]{Sample Q query}
        $S \leftarrow S \cup S^c$\\
        $Q \leftarrow Q \cup Q^c$\\
    }

    \BlankLine
    \Comment*[r]{2. Learning Metric}
    \For {$i \gets 1$ \KwTo $|S|$ }
    {
        $(x_s, y_s) \leftarrow S[i]$

        \For {$j \gets 1$ \KwTo $|Q|$ }
        {
            $(x_q, y_q) \leftarrow Q[j]$\\
            $d_{ij} \leftarrow d_{\theta_2}(g_{\theta_1}(x_s),g_{\theta_1}(x_q)))$  \Comment*[r]{Compute distances}
        }
    }
    
    \BlankLine
    Compute total loss $\mathcal{L}$ based on $d_{ij}$'s such that $d_{ij}$ is minimum when $y_i = y_j$ and maximum otherwise.
    \BlankLine
    Update parameters $\theta_1$ and $\theta_2$ on $\mathcal{L}$.  
}

\BlankLine
\textbf{Testing}: Sample a random M-way K-shot episode but this time using the classes from $N_{test}$ and evaluate its performance.

\caption{Episodic Training in Metric-based Meta-Learning Methods}
\label{alg:episodic}

\end{algorithm} 
\begin{table}[!ht]
    \centering
    \begin{tabular}{p{2cm}|p{0.5cm}|p{1.5cm}|p{1.5cm}|p{4cm}|p{2.25cm}}
    \hline
        \textbf{Method} & \textbf{T.I} & \textbf{$g_{\theta_1}$} & \textbf{$d_{\theta_2}$} & \textbf{Prediction} & \textbf{Loss} \\ \hline
        Siamese Networks \cite{Koch2015SiameseNN} & Yes & CNN &  L1  & $v = w \cdot d(g_{\theta_1}(x_1), g_{\theta_2}(x_2))$  $p = \text{sigmoid}(\sum_j v_j)$   &  $- (y \log (p) + (1-y) (\log(1-p))$ \\
        \hline
        Matching Networks \cite{Matching2016} & Yes & CNN + LSTM w/ attention & Cosine Similarity & $\hat{y} = \sum_{k=1}^t \sigma(d (f_{\theta}(\hat{x}), g_{\theta_1}(x_k)) y_k$  $P(y=c |\hat{x}) = \hat{y}_{c}$  & $-\log P$ \\
        \hline
        Prototypical Networks \cite{Snell2017PrototypicalNF} & Yes & CNN &  Euclidean & $P(y = c | x) = \sigma(-d(g_{\theta_1}(x),\mathbf{v}_c))$  & $- \log P$ \\
        \hline
        Relation Networks \cite{Sung2018LearningTC} & Yes & CNN & Learned by CNN & $r_c = d_{\theta_2}(g_{\theta_1}(x) \oplus \mathbf{v_c)})$ & $\sum\limits_{c \in C} (r_c - \mathbf{1}(y == c))^2$ \\
        \hline
        TADAM \cite{Oreshkin2018TADAMTD} & No & ResNet-12 & Cosine / Euclidean & $P_{\lambda}(y = c|x) = \sigma(-\lambda d(g_{\theta_1}(x,\Gamma),\mathbf{v}_c))$  & $-\log P$ \\
        \hline
        TapNet \cite{Yoon2019TapNetNN} & No & Resnet-12 & Euclidean &  $P(y = c|x) = \sigma(- d(\mathbf{M}(g_{\theta_1}(x)),\mathbf{M}(\Phi_c)))$  & $-\log P$ \\
        \hline
        CTM \cite{Li2019FindingTF} & No & Any  & Any & - & - \\
    \hline
    \end{tabular}
    \caption{Metric-based Meta-Learning Methods}
    \label{table:metric-ml}
\end{table}
Table \ref{table:metric-ml} compares the recent metric-based meta-learning methods based on their characteristics like embedding function, distance measure, prediction method, loss function and if the embedding function is fixed for all tasks i.e., task-independent (\textbf{T.I}) or is adaptive (task-dependent). In the following paragraphs we discuss each of these methods in further detail.

\pagebreak

\begin{figure*}[ht]
    \centering
    \includegraphics[keepaspectratio,width=0.8\textwidth]{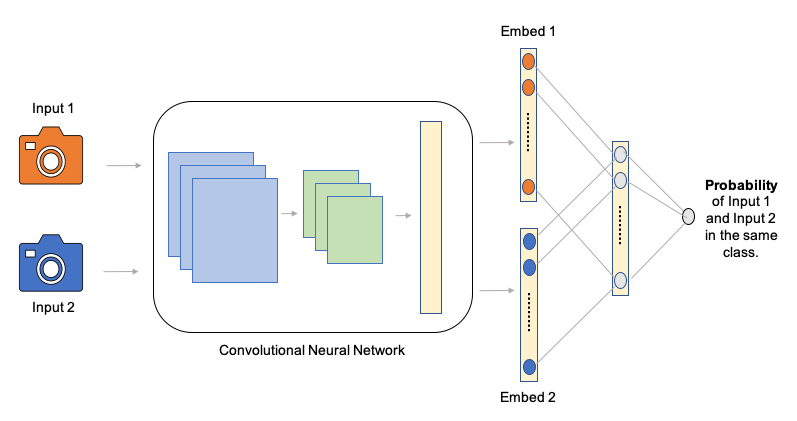}
    \caption{Convolutional Siamese Network}
    \label{fig:siamese}
\end{figure*}

\textbf{Convolutional Siamese Networks} \\
Siamese Neural Networks \cite{Bromley1993SignatureVU} are a pair of identical neural networks with shared weights originally proposed for signature verification. The two networks are joined at their output, where the joining neurons measures the distance between the feature vector output of each network. In 2015, Koch et al. \cite{Koch2015SiameseNN} used a pair of identical convolutional neural networks (CNNs) \cite{Lecun98gradient-basedlearning} with shared weights as done in Siamese Networks for image verification. The network was trained to recognize if the two images belong to the same class or not. The network outputs a probability score (similarity) of the two images belonging to the same class. This idea was further extended to perform one-shot recognition by comparing the similarity score between a query image and support images from different classes. This is illustrated in Figure \ref{fig:siamese} . Here the embedding function $g_{\theta_1}$ is a CNN and the distance between two embedding vectors is simply the L1 distance i.e., $|g_{\theta_1}(x_1) - g_{\theta_1}(x_2)|$. The distance is converted to the probability of similarity by a linear feed-forward layer and a sigmoid function. The network is then trained on binary cross-entropy loss. Similar to Convolutional Siamese Networks, Mehrotra et al. \cite{MehrotraD19} proposed Skip Residual Pairwise Networks (SRPNs) where they used a Wide Residual Network \cite{BMVC2016_87} as the embedding function $g_{\theta_1}$  and a network of residual blocks to act as the distance function  $d_{\theta_2}$.

\begin{figure*}[ht]
    \centering
    \includegraphics[keepaspectratio,width=0.6\textwidth]{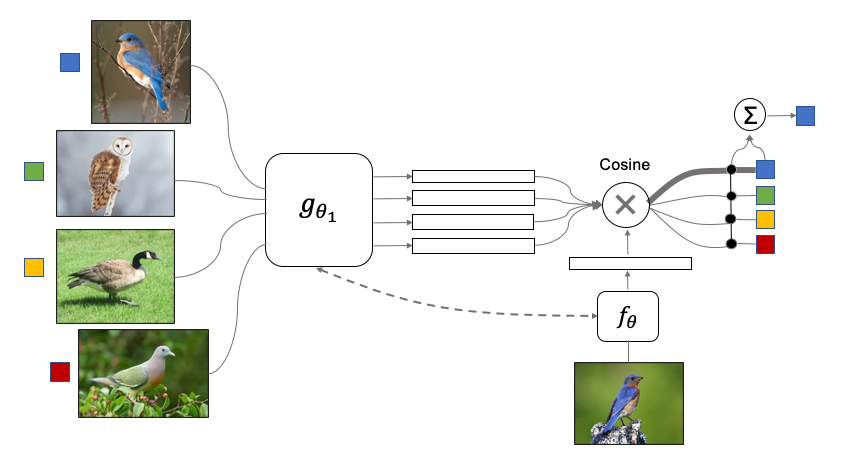}
    \caption{Matching Networks (Figure adapted from \cite{Matching2016}).}
    \label{fig:matching}
\end{figure*}

\textbf{Matching Networks} \\
Given a support set $S = \{x_i, y_i\}_{k=1}^K$ and a query $\hat{x}$,  Matching Networks \cite{Matching2016} (Figure \ref{fig:matching}) define a probability distribution over the output labels $y$ using an attention kernel $a(\hat{x}, x_k)$ . The attention kernel basically computes the cosine similarity between the embeddings of support and query examples and then normalize the similarity score by taking a softmax:

\begin{equation} \label{eq:matching}
    a(\hat{x}, x_k) = e^{\cos(f(\hat{x}),g(x_k))} / \sum_{k=1}^t e^{\cos(f(\hat{x}),g(x_k))}. 
\end{equation}

The classifier's output is then defined as a sum of the labels (one-hot encoded) of support samples weighted by the attention kernel $a(\hat{x}, x_k)$ :

\begin{equation} \label{eq:matching-output}
    P(y|\hat{x},S) = \sum_{k=1}^K a(\hat{x}, x_k) y_k
\end{equation}

In a simple case, the embedding function for query ($f_\theta$) and the embedding function for examples in the support set ($g_{\theta_1}$) are same i.e., $f = g$.  Alternatively, Matching Networks propose using Full Context Embeddings where embedding functions contextually embeds images i.e for a given support image $x$ and support set $S$, the embedding of $x$ is obtained by $g_{\theta_1}$ in the presence of $S$ as $g_{\theta_1}(x;S)$ . Similarly for a query $\hat{x}$, its embedding would be $f_{\theta}(\hat{x};S)$.  i.e., $S$ should be able to modify how we embed $\hat{x}$ through $f$.  Using contextual embedding showed in improvement in the performance of few-shot classification on miniImageNet but no difference was observed on the simpler omniglot dataset.

\textbf{Prototypical Networks} \\
Prototypical Networks \cite{Snell2017PrototypicalNF} use a 4-layer-CNN as an embedding function $g_{\theta_1}$. A prototype for each class is defined by taking an average of embedding vectors obtained from the support images belonging to that class (Eqn. \ref{eq:prototype}):

\begin{equation} \label{eq:prototype} 
    \mathbf{v}_c  = \frac{1}{|S^c|} \sum_{(x_k,y_k) \in S^c} g_{\theta_1}(x_k).
\end{equation}

The similarity is measured by calculating the squared euclidean distance between the query's embedding and each class prototype. The output probability over classes is calculated by taking a softmax over the negative distances (Eqn. \ref{eq:proto-softmax}):

\begin{equation} \label{eq:proto-softmax}
  P(y = c|\hat{x}) = \text{softmax}(-d(g_{\theta_1}(\hat{x}), \mathbf{v}_c)) = \frac{e^{(-d(g_{\theta_1}(\hat{x}),\mathbf{v}_c))}} {\sum_{\hat{c} \in C} e^{(-d(g_{\theta_1}(\hat{x}),\mathbf{v}_{\hat{c}}))}}.  
\end{equation}

The loss $\mathcal{L}$ is given by the negative log-likelihood of the correct class (Eqn. \ref{eq:proto-loss}):

\begin{equation} \label{eq:proto-loss}
  \mathcal{L}(\theta_1) = - \log P_{\theta_1}(y = c | \hat{x}).  
\end{equation}

Furthermore, to generate more discriminative embeddings, Zhang et al. \cite{Zhang2019OneSL} proposed using a multi-way contrastive loss function with margin to pull the examples belonging to the same class together and push the others away in the embedding space. 

\begin{figure*}[ht]
    \centering
    \includegraphics[keepaspectratio,width=0.8\textwidth]{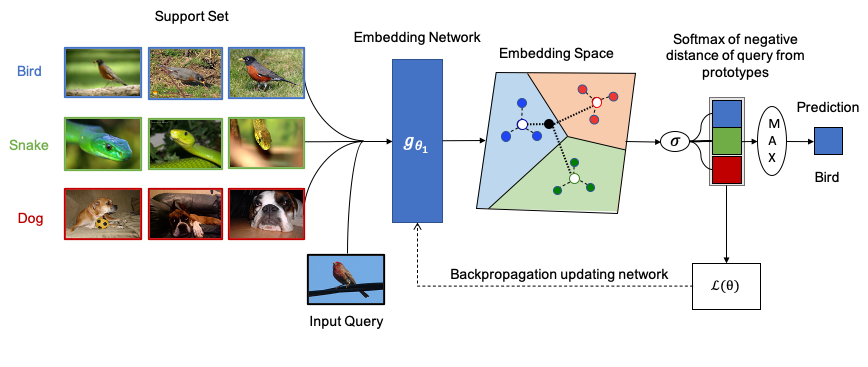}
    \caption{Few-shot prototypes $\mathbf{v_c}$ are computed as the mean of embedded support examples for each class. The embedded query points are classified via a softmax over distances to the class prototypes.}
    \label{fig:proto}
\end{figure*}

\textbf{Relation Networks} \\
Relation Networks \cite{Sung2018LearningTC} get rid of manually constructing a similarity metric and instead just employs another CNN to output a similarity score. The representations of the support and query examples (obtained from $g$) are concatenated together and fed into $d$ to obtain a relation score between each query and a support class (Figure \ref{fig:metric}). When the number of support examples per class are more than one, the relation score is calculated between a query and a prototype of a class $\mathbf{v_c}$, which is an element wise sum of embeddings of support examples of that class. The relation score for a match should be 1 and 0 for a no-match. The network is then trained to minimize the mean squared error of relation scores.

\textit{Relation score of query $\hat{x}$ with class $c$ :}

$$ 
r_c = d_{\theta_2}(g_{\theta_1}(\hat{x}) \oplus \mathbf{v_c}) 
$$

\textit{Optimization Objective:}

 $$
 \theta_1, \theta_2 \leftarrow \text{arg}\min_{\theta_1, \theta_2}\sum\limits_{c \in C} (r_c - \mathbf{1}(y == c))^2
 $$

\begin{figure*}[ht]
    \centering
    \includegraphics[keepaspectratio,width=0.7\textwidth]{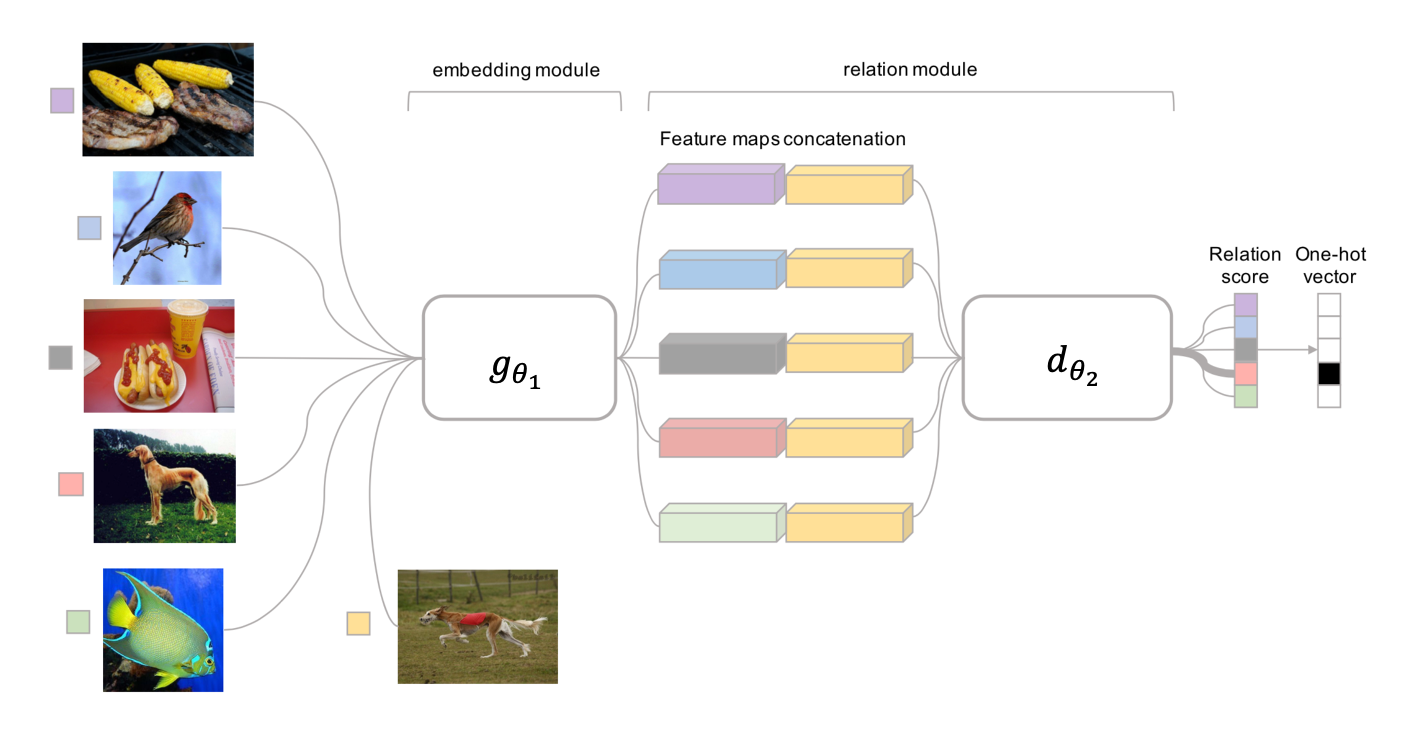}
    \caption{Relation Network (Source: \cite{Sung2018LearningTC})}
    \label{fig:relation}
\end{figure*}

\textbf{Task-Dependent Adaptive Metric (TADAM)} \\
In the approaches discussed previously, the embedding function $g_{\theta_1}$ was \textit{task-independent}, meaning given any task $\mathcal{T} \sim p(\mathcal{T})$, its samples will be embedded using a fixed embedding function   $g_{\theta_1}$. Moreover, the choice of distance metric function (i.e. cosine or euclidean) was also something to be experimented with depending upon the task at hand.
Oreshkin et al. \cite{Oreshkin2018TADAMTD} proposed using \textbf{1)} a learnable softmax temperature \cite{Hinton2015DistillingTK} with a scaling factor $\alpha$ to bridge the gap between the performance of cosine and euclidean distance metric and \textbf{2)} a Task Embedding Network (TEN) factored into the embedding network $g_{\theta_1}$ to output task adaptive representations. These contributions combined with using a deeper embedding network (Resnet-12 \cite{he2016deep}) as the feature extractor resulted in 8.5\% absolute accuracy improvement over Prototypical Networks \cite{Snell2017PrototypicalNF} on the \textit{mini}ImageNet \cite{Matching2016} 5-way 5-shot classification task.

\begin{enumerate}
    \item \textbf{Metric Scaling:}  It was observed that Prototypical Networks \cite{Snell2017PrototypicalNF} which used euclidean distance performed better on few-shot image classification when compared to Matching Networks \cite{Matching2016} which used cosine distance. Oreshkin et al. \cite{Oreshkin2018TADAMTD} suggested that the difference in performance could be directly attributed to the interaction of the different scaling of the metrics with the softmax. Hence they propose to scale the distance metric by a learnable temperature, $\lambda$, and observed that both distances produced equivalent performance when scaled with learned parameter $\lambda$ (Eqn. \ref{eq:softmax-temp}):
    
    \begin{equation}  \label{eq:softmax-temp}
        P_{\lambda}(y = c|x) = \text{softmax}(-\lambda d(g_{\theta_1}(x),\mathbf{v}_c)). 
    \end{equation}
    
    \item \textbf{Task Conditioning:}  Previously, Matching Networks used contextual embeddings i.e, embedding for an input image $x$ was obtained in presence of its support set $S$, $g_{\theta_1}(x;S)$. This was achieved with a bidirectional LSTM as a post-processing of a fixed feature extractor. Differently, TADAM explicitly define a dynamic feature extractor $g_{\theta_1}(x, \Gamma)$ where $\Gamma$ is the set of parameters predicted from a task representation such that the performance of $g_{\theta_1}(x, \Gamma)$ is optimized given the task sample set $S$.

\end{enumerate}

\textbf{Task-Adaptive Projection (TapNet)} \\
In addition to the embedding function $g_{\theta_1}$ and the distance function $d$, TapNet \cite{Yoon2019TapNetNN} proposed a concept of per-class reference vectors $\mathbf{\Phi}$ and a task dependent projection space or mapping $\mathbf{M}$. Unlike class prototypes in Prototypical Networks, reference vectors $\Phi$ for each class are learned. The projection space $\mathbf{M}$ is non-parametric and is built specific to each task. The input query $x$ is then classified by measuring its distance to different reference vectors $\Phi$ in the projection space $\mathbf{M}$ (Eqn. \ref{eq:projection}):

\begin{equation} \label{eq:projection}
    P(y = c|x) = \text{softmax}(- d(\mathbf{M}(g_{\theta_1}(x)),\mathbf{M}(\Phi_c))).
\end{equation}

The motivation is to find a projection space $\mathbf{M}$ which could remove the misalignment between the task-embedded features and the references and thus resulting in better classification performance.

\textbf{Task-Relevant Features (CTM)} \\
Similar to Matching Networks, Li et al. \cite{Li2019FindingTF} propose using a plugable component called Category Traversal Module (CTM) parameterized by $\phi$ to find contextual embeddings of the images in the support set $S$ and the query set $Q$.  The parameters $\phi$ are learned during the training along with parameters $\theta_1$ of the embedding function $g$. 

$$\text{CTM}(g_{\theta_1}(S);\phi) \rightarrow I(S)$$
$$\text{CTM}(g_{\theta_1}(Q);\phi) \rightarrow I(Q)$$

\begin{figure*}[ht]
    \centering
    \includegraphics[keepaspectratio,width=0.8\textwidth]{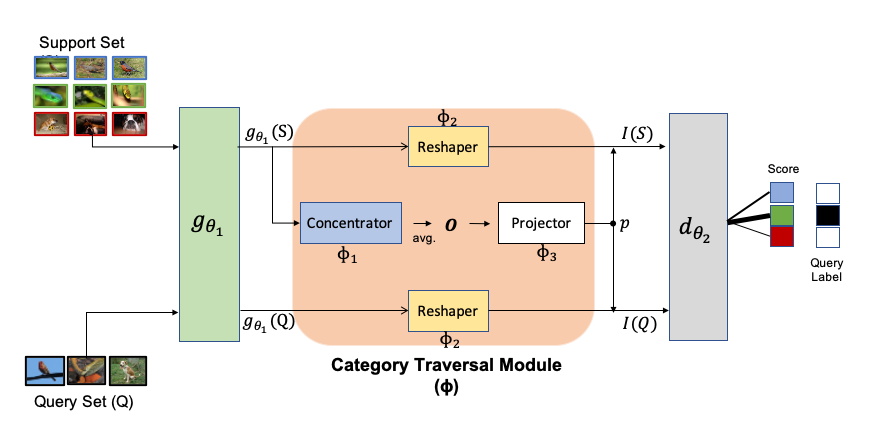}
    \caption{Category Traversal Module (CTM) (Figure adapted from \cite{Li2019FindingTF})}
    \label{fig:CTM}
\end{figure*}

The CTM takes support set features $g_{\theta_1}(S)$ as an input and produces a mask $p$ via a concentrator and projector that make use of intra and inter-class views respectively. The mask $p$ is applied to reduced-dimension features of both the support and query, producing improved features $I$ with dimensions relevant for the current task. These improved feature embeddings are finally fed into a metric learner.  This is illustrated in Figure \ref{fig:CTM}.

\subparagraph{Attention-based methods} \mbox{} \\
Much like Matching Networks and CTM, few others have proposed integrating attention \cite{Vaswani2017AttentionIA} modules in existing methods for learning more discriminative feature embeddings. Hou et al. \cite{Hou2019CrossAN} proposed Cross Attention Networks (CAN) with a Cross Attention Module (CAM) to transform the class prototypes $P$ and query $Q$ to more discriminative prototypes $\bar{P}$ and query $\bar{Q}$. Furthermore, Hao et al. \cite{Hao2019CollectAS} propose Semantic Alignment Metric Learning (SAML) method to align semantically relevant local regions on support and query images using attention maps.

\pagebreak

\paragraph{Optimization-based Meta-Learning} \label{sec:optim} \mbox{}\\
Earlier in Section \ref{sec:fsc}, we discussed that for a few-shot classification task $\mathcal{T}$ with training data $\mathcal{D}^{train}$ (Eqn. \ref{eq:dtrain}) where the number of training examples $t$ is small, it is difficult to approximate $f$ (Eqn. \ref{eq:simple-predict}) with parameters $\theta$ (Eqn. \ref{eq:simple-theta}) from scratch using gradient-based optimization as its not designed to cope with small number of training samples and thus will lead to overfitting. This ponders the question, "is there any way to optimize on limited training data and still achieve good generalization performance?" Optimization-based meta-learning for FSL answers to this question. Basically, leveraging the meta-learning architecture (Figure \ref{fig:meta-setup}) and episodic training (Algorithm \ref{alg:episodic}), optimization-based methods enables an optimization procedure to work on limited training examples.   

\textbf{Learner and Meta-Learner} 
\newline
Optimization-based methods generally involves learning in two stages:

\begin{enumerate}
    \item \textbf{Learner:}  A learner model $f_{\theta}$ is task-specific and trained for a given task. For a given few-shot task, a stand-alone learner model trained from scratch using gradient descent (Eqn. \ref{eq:simple-theta}) will not be able to generalize (Eqn. \ref{eq:simple-predict}).   
    
    \item \textbf{Meta-Learner:}  A meta-learner model $g_{\phi}$ is not task specific and is trained on a distribution of tasks $\mathcal{T} \sim p(\mathcal{T})$ (Figure \ref{fig:meta-setup}). Using episodic training, the meta-learner learns ($\phi$) to update the learner model's parameters ($\theta$) via training set $\mathcal{D}^{train}$, 
        \begin{equation} \label{eq:meta-update} 
            {\theta^*} = g_{\phi}(\theta, \mathcal{D}^{train}).  
        \end{equation}
    The objective of the meta-learner model is to produce updated learner model parameters $\theta^*$ such that they are better than stand-alone learner model parameters $\theta$. 
    
\end{enumerate}

During meta-training (notation \textbf{A} in Table \ref{table:nom}), the optimization process involves updating $\phi$ for the meta-learner and $\theta$ for individual training tasks. Once the meta-training finishes, the prior knowledge is encompassed into $\phi$ and only $\theta$ is updated for a test task (Eqn. \ref{eq:meta-update}). 

Table \ref{table:optim} compares different optimization-based meta-learning methods based on how they update learner's parameters $\theta$ and meta-learner parameters $\phi$. In all the listed methods, learning happens in two-stages. Initially, in the outer-loop, the meta-learner's parameters $\phi$ are randomly initialized. Next, in the inner-loop the learner parameters ($\theta$) are updated/proposed by meta-learner (Eqn. \ref{eq:meta-update}). The learners' training loss $\mathcal{L}^{train}$ is further used to obtain optimal parameters $\theta^*$. Finally, in the outer loop the cumulative test loss of learner obtained using $\theta^*$ is used for updating $\phi$. In some cases learner's initial parameters $\theta$ are also meta-learned along with $\phi$.

In the following paragraphs, we discuss the the recent optimization based meta-learning approaches in more detail.

\textbf{LSTM Meta-Learner} \\
Normally when given a task $\mathcal{T}$, we try to learn function $f(\theta)$ on its training data $\mathcal{D}^{train}$. The parameters $\theta$ of the neural network $f$ are updated using some form of gradient descent as such:

\begin{equation} \label{eq:theta-update}
    \theta_{i+1} = \theta_i - \alpha \nabla f(\theta_i).
\end{equation}

Instead of using an hand designed optimizer such as SGD and a fixed learning rate $\alpha$, Andrychowicz et al. \cite{Andrychowicz2016LearningTL} learn an optimizer function $g_{\phi}$ such that:

\begin{equation} \label{eq:update-predict}
  \theta_{i+1} = \theta_i + g_i(\nabla f(\theta_i); \phi).  
\end{equation}

Similarly, Ravi \& Larochelle \cite{Ravi2017OptimizationAA} modeled an LSTM \cite{Hochreiter1997LongSM} as a meta-learner $g_{\phi}$ to propose parameters for the learner $f$:

\begin{equation} \label{eq:theta-predict}
  \theta_{i+1} = g_i(\nabla f(\theta_i), \theta_i; \phi).  
\end{equation}

\begin{figure*}[ht]
    \centering
    \includegraphics[keepaspectratio,width=0.8\textwidth]{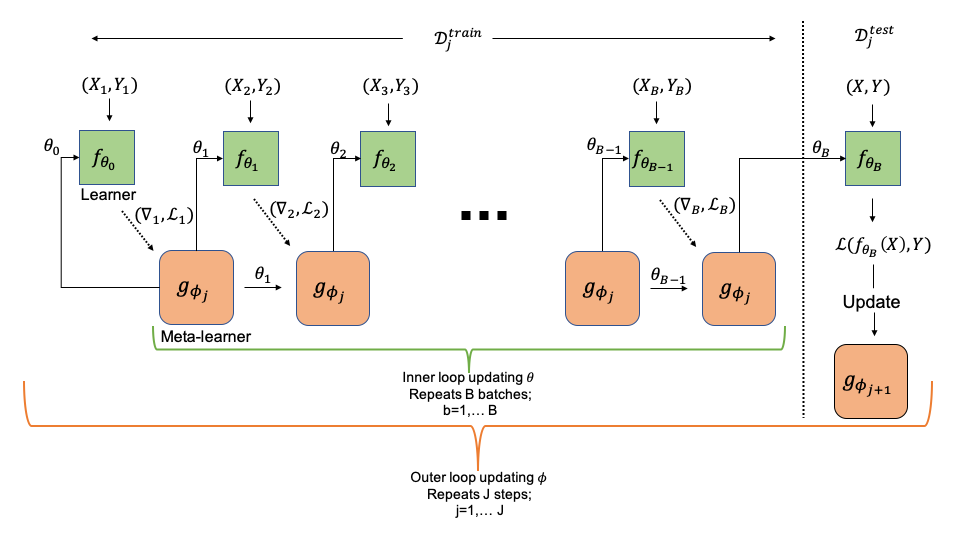}
    \caption{Computational graph for the forward pass of the meta-learner (Figure adapted from \cite{Ravi2017OptimizationAA}).}
    \label{fig:lstm-ml}
\end{figure*}

Optimizer $g$ is trained to produce parameters $\theta$ in just few steps for the few-shot learning task $\mathcal{T}$. It follows the episodic training paradigm and mimics the testing scenario. This training procedure is described in Algorithm \ref{alg:LSTM} and depicted in Figure \ref{fig:lstm-ml}.

\begin{table}[]
    \centering
    \begin{tabular}{lp{5.5cm}p{5cm}}
        \hline
        \textbf{Method} & \textbf{Learner}  & \textbf{Meta-Learner} \\ \hline \\
     
            &   Repeat $\forall b \in [1..B]$   &   Repeat $\forall j \in [1..J]$ \\
        
        LSTM Meta-Learner \cite{Ravi2017OptimizationAA} &   $$\mathcal{L}_b \leftarrow \mathcal{L}(f(X_{b};\theta_{b-1}), Y_b)$$ $$\theta_b \leftarrow g((\nabla_{\theta_{b-1}} \mathcal{L}_b, \mathcal{L}_b);\phi_{j-1})$$  &   $$\mathcal{L}^{test}_j \leftarrow \mathcal{L}(f(X;\theta_{B}), Y)$$ $$\phi_{j} \leftarrow \phi_{j-1} - \alpha \nabla_{\phi_{j-1}}\mathcal{L}^{test}_j$$ \\
        
        \hline
        \\
            &  Repeat $\forall i \in [1..I]$ &  Repeat $\forall j \in [1..J]$ \\
    
    MAML \cite{MAML2017} &  $$\mathcal{L}_i^{train} \leftarrow \mathcal{L}(f(\mathcal{D}^{train}_i;\theta_{j-1}))$$ $$\theta_i^* \leftarrow \theta_{j-1} - \alpha \nabla_{\theta_{j-1}}\mathcal{L}_t^{train}$$    $$\mathcal{L}_i^{test} \leftarrow \mathcal{L}(f(\mathcal{D}^{test}_i;\theta_t^*))$$ & $$\theta_j \leftarrow \theta_{j-1} - \beta \nabla_{\theta_{j-1}} \sum_{i=1}^I \mathcal{L}_i^{test}$$ \\
    
    \hdashline \\

    MTL \cite{Sun2018MetaTransferLF} &  $$\mathcal{L}_i^{train} \leftarrow \mathcal{L}(f(\mathcal{D}^{train}_i;[\theta_{j-1},\phi_{j-1},\Theta]))$$ $$\theta_i^* \leftarrow \theta_{j-1} - \alpha \nabla_{\theta_{j-1}}\mathcal{L}_i^{train}$$    $$\mathcal{L}_i^{test} \leftarrow \mathcal{L}(f(\mathcal{D}^{test}_i;\theta_i^*))$$ & $$\theta_j \leftarrow \theta_{j-1} - \beta \nabla_{\theta_{j-1}} \sum_{i=1}^I \mathcal{L}_i^{test}$$  $$\phi_j \leftarrow \phi_{j-1} - \beta \nabla_{\phi_{j-1}} \sum_{i=1}^I \mathcal{L}_i^{test}$$ \\
    
    \hdashline \\
    
    LEO \cite{Rusu2018MetaLearningWL} & $$\phi_{j-1} = \{\phi_e,\phi_r,\phi_d,\alpha\}$$ $$\mathbf{z_i} \leftarrow g(\mathcal{D}^{train}_i;[\phi_e,\phi_r,\Theta])$$ $$\theta_i \leftarrow g(\mathbf{z_i}; \phi_d)$$  $$ \mathcal{L}^{train}_i \leftarrow \mathcal{L}(f(\mathcal{D}^{train}_i;\theta_i)) $$ $$ \mathbf{z_i^*} \leftarrow \mathbf{z_i} - \alpha \nabla_{\mathbf{z_i}}\mathcal{L}_i^{train} $$ $$\theta_i^* \leftarrow g(\mathbf{z_i^*}; \phi_d)$$ $$ \mathcal{L}^{test}_i \leftarrow \mathcal{L}(f(\mathcal{D}^{test}_i;\theta_i^*))$$ & $$\phi_j \leftarrow \phi_{j-1} - \beta \nabla_{\phi_{j-1}} \sum_{i=1}^I \mathcal{L}_i^{test}$$ \\
    \hline
    \end{tabular}
    \caption{Optimization-based Meta-Learning Methods}
    \label{table:optim}
\end{table}

\begin{algorithm}

\DontPrintSemicolon
\SetAlgoLined

\textbf{Input: }
Meta-training set $D_{meta-train}$, Learner $f$ with parameters $\theta$, Meta-Learner $g$ with parameters $\phi$
\BlankLine

$\phi_{0} \leftarrow $ random initialization

\For {$j \gets 1$ \KwTo $J$ }
{
    $D^{train}, D^{test} \leftarrow $ random dataset from $D_{meta-train}$\\
    $\theta_0 \leftarrow c_0$
    \BlankLine
    
    \For{$b \gets 1$ \KwTo $B$}
    {
        $\mathbf{X_b}, \mathbf{Y_b} \leftarrow $ random batch from $D^{train}$\\
        $\mathcal{L}_b \leftarrow \mathcal{L}(f(\mathbf{X_b};\theta_{b-1}),\mathbf{Y_b})$  \Comment*[r]{Get loss of learner on train batch}
        $c_b \leftarrow g((\nabla_{\theta_{b-1}} \mathcal{L}_b, \mathcal{L}_b);\phi_{j-1})$ \Comment*[r]{Get output of meta-learner}
        $\theta_b \leftarrow c_b$ \Comment*[r]{Update learner parameters}
    }
    \BlankLine
    $\mathbf{X, Y} \leftarrow D^{test}$\\
    $\mathcal{L}_{test} \leftarrow \mathcal{L}(f(\mathbf{X};\theta_b), \mathbf{Y})$ \Comment*[r]{Get loss of learner on test batch}
    Update $\phi_j$ using $\nabla_{\phi_{j-1}}\mathcal{L}_{test}$ \Comment*[r]{Update meta-learner parameters}
    
}

\caption{Train Meta-Learner}
\label{alg:LSTM}
\end{algorithm}

\pagebreak

\begin{figure*}[ht]
    \centering
    \includegraphics[keepaspectratio,width=0.5\textwidth]{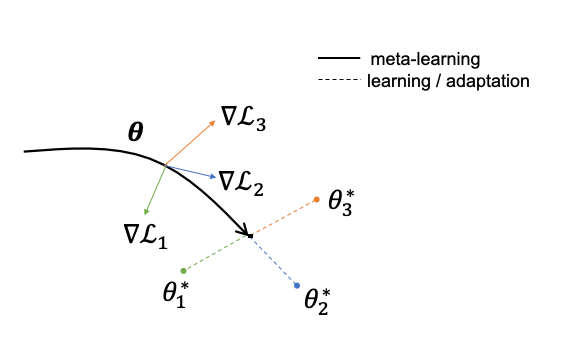}
    \caption{Diagram of MAML, which optimizes for a representation $\theta$ that can quickly adapt to new tasks (Figure adapted from \cite{MAML2017}).}
    \label{fig:MAML}
\end{figure*}

\textbf{Model-Agnostic Meta-Learning (MAML)} \\
They key idea in MAML or Model-Agnostic Meta-Learning \cite{MAML2017} is to achieve good initialization parameters $\theta$ such that new tasks can optimize quickly from $\theta$ through one or more gradient descent steps computed with a small amount of data (Figure \ref{fig:MAML}). These initial parameters $\theta$ are meta-learned over a distribution of tasks $p(\mathcal{T})$. Unlike LSTM meta-learner which has two separate models, a meta-learner model $g_{\phi}$ and a task learner model $f_{\theta}$, MAML has a single model with parameters $\theta$. Individual tasks use the model parameters $\theta$ as initialization to arrive at optimal parameters $\theta^*$ for the task at hand (Algorithm \ref{alg:MAML}). 

\begin{algorithm}

\DontPrintSemicolon
\SetAlgoLined

\textbf{Require:} $p(\mathcal{T}):$ distribution over tasks\\
\textbf{Require:} $\alpha, \beta:$ step size hyperparameters

$\theta_{0} \leftarrow $ random initialization

\For {$j \gets 1$ \KwTo $J$ }
{
    Sample a batch of $I$ tasks randomly from $p(\mathcal{T})$\\
    \For {$i \gets 1$ \KwTo $I$}
    {
        $D^{train}_i, D^{test}_i \leftarrow $ dataset for task $\mathcal{T}_i$ from $D_{meta-train}$\\    
        $\mathcal{L}_i^{train} \leftarrow \mathcal{L}(f(D^{train}_i;\theta_{j-1}))$ \Comment*[r]{Get loss of learner on training data}
        $\theta_i^* \leftarrow \theta_{j-1} - \alpha \nabla_{\theta_{j-1}}\mathcal{L}_i^{train}$ \Comment*[r]{Adapt to learner's training loss}
        $\mathcal{L}_i^{test} \leftarrow \mathcal{L}(f(D^{test}_i;\theta_i^*))$ \Comment*[r]{Get test loss of learner on adapted parameters}
        
    }

    $\theta_j \leftarrow \theta_{j-1} - \beta \nabla_{\theta_{j-1}} \sum\limits_{i=1}^I \mathcal{L}_i^{test}$ \Comment*[r]{Update parameters on total test loss of learners}
    \BlankLine
}

\caption{MAML}
\label{alg:MAML}

\end{algorithm}

\textit{Proto-MAML} \cite{Triantafillou2020MetaDatasetAD} extends MAML by combining ideas from Prototypical Networks \cite{Snell2017PrototypicalNF} and MAML \cite{MAML2017}. During training, the former focuses on learning a good embedding function $g_{\theta}$ and does not perform any task adaptation whereas later approach focuses on learning a good initialization parameters $\theta$ which are then fine-tuned to adapt to each task. In Proto-MAML, the initial weights of the classifier are obtained from prototypical networks which are then subsequently fine-tuned to the individual task.

\textit{Task Agnostic Meta-Learning (TAML)} \cite{Jamal2018TaskAM} argues that the initial model of the meta-learner could be too biased toward existing tasks to adapt to new tasks, especially in the case of few-shot learning with discrepancy between new tasks from those in the training tasks. In  such case, we wish to avoid an initial model over-performing  on some tasks. Therefore, TAML aims to meta-train an unbiased initial model by preventing it from over-performing on some tasks or directly minimizing the inequality of performance across different tasks, in hope to make it more generalizable to unseen tasks.  

Antoniou et al. \cite{Antoniou2018HowTT} argue that even though MAML is a simple yet elegant meta-learning framework, it suffers from a variety of problems  which can cause 1) instability during training, 2) restricted generalization performance, 3) reduction in the framework's flexibility, 4) increase in the system's computational overhead and 5) a costly hyperparameter tuning before it can work robustly on a new task. Hence they proposed \textit{MAML++} \cite{Antoniou2018HowTT}, an improved meta-learning framework that offers the flexibility of MAML along with stability, computational efficiency and improved generalization performance.

\begin{figure*}[ht]
    \centering
    \includegraphics[width=0.8\textwidth]{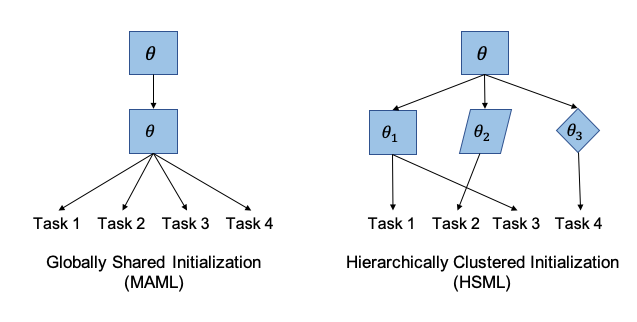}
    \caption{MAML vs HSML (Figure adapted from \cite{Yao2019HierarchicallySM})}
    \label{fig:HSML}
\end{figure*}

\textit{Hierarchically Structured Meta-Learning (HSML) }\cite{Yao2019HierarchicallySM} attempts to address the challenge of task uncertainty and heterogeneity in meta-learning, which can not be handled by globally sharing knowledge among tasks or to learn a single initialization for all kinds of tasks as done in MAML. Therefore, HSML explicitly tailor transferable knowledge to different clusters of tasks by learning task representations where each cluster of tasks has its own initial parameters (Figure \ref{fig:HSML}).  

\textit{Fast \textbf{C}ontext \textbf{A}daptation \textbf{Via} Meta-Learning} (CAVIA) \cite{Zintgraf2019FastCA} is a simple extension to MAML that is more interpretable and less prone to overfitting. CAVIA partitions the model parameters $\theta$ into two parts: context parameters $\theta_{context}$ that serve as additional input to the model and are adapted on individual tasks, and shared parameters $\theta_{shared}$ that are meta-trained and shared across the tasks. At test time, only context parameters are updated, leading to a low dimensional task representation. 

\begin{figure*}[ht]
\centering
\includegraphics[keepaspectratio,width=0.8\textwidth]{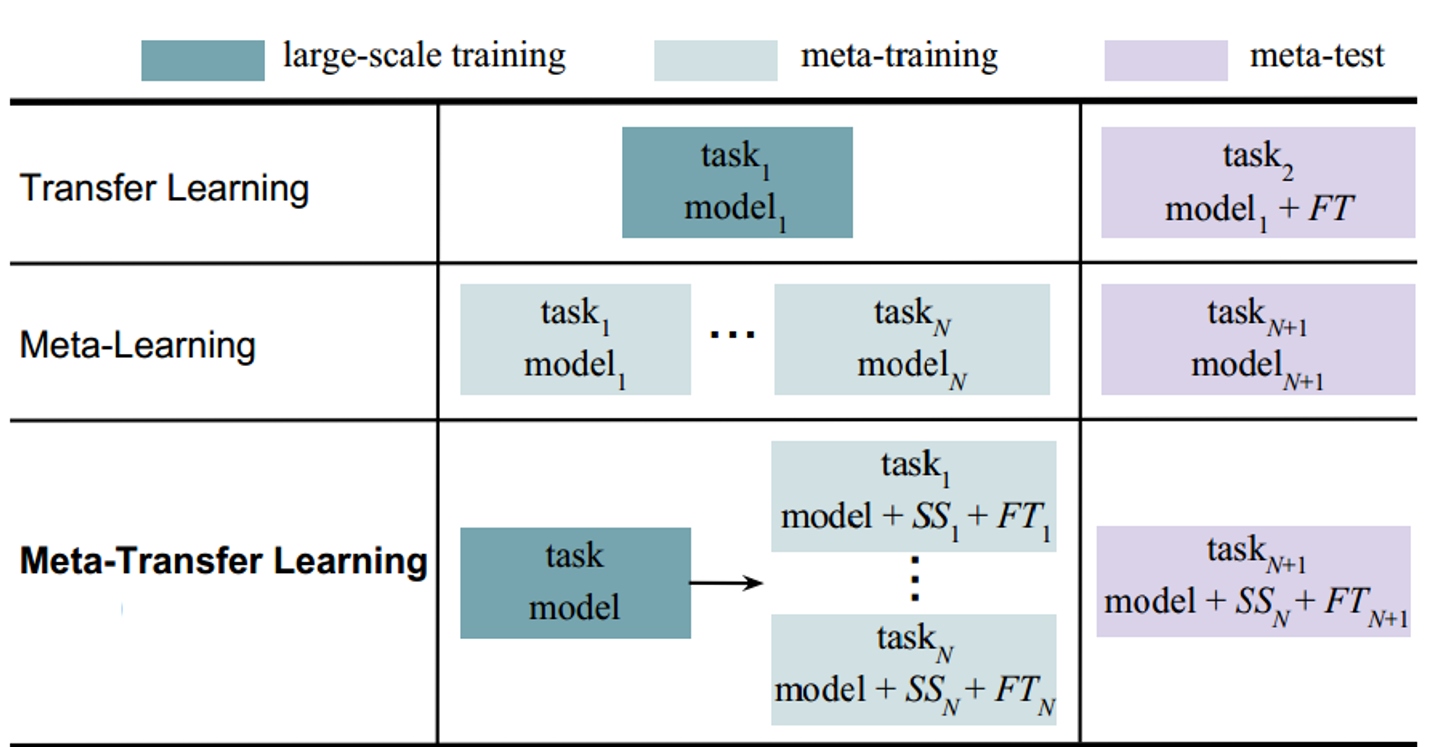}
\caption{Difference between learning strategies. Meta-Task Learning (MTL) adopts transfer learning strategy for Meta-Learning. A pre-trained model’s weights are shift and scaled  on task basis. The shift and scale (SS) parameters are learned through meta-learning across tasks (Source: \cite{Sung2018LearningTC}).}
\label{fig:MTL}
\end{figure*}

\pagebreak
\textbf{Meta-Transfer Learning (MTL)} \\   
In MAML \cite{MAML2017}, the meta-learned model parameters $\theta$ are adapted as $\theta^*$ to an individual task. The effectiveness of this strategy is limited to shallow networks, as the adaptation in deep networks can lead to overfitting. Therefore, Sun et al. \cite{Sun2018MetaTransferLF} propose a technique called Meta-Transfer Learning (MTL) (Figure \ref{fig:MTL}). The key idea is to use a pretrained DNN as a feature extractor ($\Theta$) and meta-learn only the last layer classifier parameters $\theta$.  In addition, MTL also meta-learns scale ($\phi_{S_1}$) and shift ($\phi_{S_2}$) parameters to adapt the $\Theta$ to individual tasks. The scale and shift parameters are few in number when compared to $\Theta$.

\pagebreak

\textbf{Latent Embedding Optimization} \\
Latent Embedding Optimization(LEO) \cite{Rusu2018MetaLearningWL} learns a low-dimensional latent embedding of model parameters and performs optimization-based meta-learning in this space. Learning low-dimensional latent representation is motivated by the fact that it is difficult to optimize in high-dimensional spaces in extreme low-data regimes. The following steps explain the optimization process of LEO:

\begin{enumerate}
    \item \textbf{Pretraining}: Like MTL, input feature embeddings are not learned, but are pretrained on a deep network (WRN \cite{BMVC2016_87}). Using the meta-training dataset, a ResNet classifier is trained to distinguish between training classes. Features from an intermediate layer of this trained classifier is used for getting input embeddings.
    \item \textbf{Inner Loop Training}: Each task's dataset $\mathcal{D}^{train}$ is used to obtain the initial parameters $\theta$ of the classifier $f$. 
    \begin{itemize}
        \item The pre-trained embeddings for examples $\mathcal{D}^{train}$ are fed into a Encoder-Relation network which outputs latent representation $\mathbf{z}$ for each class (like a prototype). 
        \item The low dimensional latent representation generates task classifier parameters $\theta$ through a decoder.
        \item  The training loss $\mathcal{L}^{train}$ of $f_{\theta}$ is used to update the latent representations $\mathbf{z}$.
    \end{itemize}
    \item \textbf{ Outer Loop Training}: The loss $\mathcal{L}^{test}$ calculated on each task's test examples $\mathcal{D}^{test}$ with $f_{\theta}$ is used to update the parameters of encoder, relation network and decoder.
\end{enumerate}

\begin{figure*}[hb]
    \centering
    \includegraphics[width=0.8\textwidth]{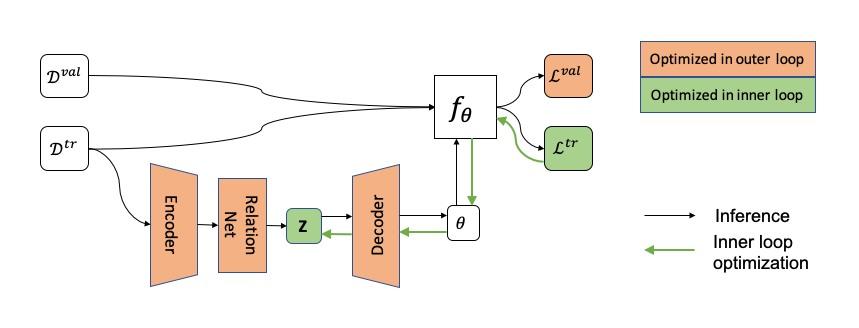}
    \caption{Overview of the architecture of LEO (Source: \cite{Rusu2018MetaLearningWL})}
    \label{fig:LEO}
\end{figure*}

Intuitively, this provides two advantages. First, the initial parameters for a new task are conditioned on the training data, which enables a task-specific starting point for adaptation. Second, by optimizing in the lower-dimensional latent space, the approach can adapt the behavior of the model more effectively (Figure \ref{fig:LEO}).

\vfill
\pagebreak

\paragraph{Model-based Meta-Learning} \mbox{}\\
Metric-based methods for FSL learn a function $g_{\theta}$ to generate discriminative embeddings based on a metric where as optimization-based methods learn a priors {$\phi,\theta$} to optimize quickly from. Different from these approaches, model-based meta-learning methods makes no assumption on the form of $P_{\theta}(y | x)$. Rather it involves model architectures specifically tailored for fast learning. Table \ref{table:model-based} summarizes the methods in this category. Based on the kind of model architecture ($f_{\theta}$), these methods are further categorized into memory-based, rapid-adaptation-based and miscellaneous models.

\begin{table}[!ht]
    \centering
    \begin{tabular}{l|l|l}
    \hline
         \textbf{Method} & \textbf{Type} & \textbf{Key Idea} \\ \hline
         MANN \cite{Santoro2016OneshotLW} & Memory & Using NTM \cite{Graves2014NeuralTM} for sequence prediction \\
         MM-Net \cite{Cai2018MemoryMN} & Memory & Key-Value Memory Networks \cite{Miller2016KeyValueMN} combined with Matching Networks \cite{Matching2016} \\
         MetaNets \cite{Munkhdalai2017MetaN} & Rapid Adaptation & Fast-Weights stored in memory \\
         CSN \cite{Munkhdalai2017RapidAW} & Rapid Adaptation & Task specific conditioning of activation values stored in memory \\
         SNAIL \cite{Mishra2017ASN} & Miscellaneous & Temporal Convolutions and Causal Attention layers for sequence prediction \\
    \hline
    \end{tabular}
    \caption{Model-based Meta-Learning Methods}
    \label{table:model-based}
\end{table}

\subparagraph{Memory as a component} \mbox{} \\
A family of model architectures integrates an external memory component to facilitate their learning process. This external memory component is usually a 2D matrix called the memory bank, memory matrix or just plain \textit{memory}. The memory acts as a storage buffer to which neural networks can write new information and retrieve previously stored information. Note that this memory component is different than \textit{internal memory} found in vanilla RNNs or LSTMs. Neural Turing Machines (NTM) \cite{Graves2014NeuralTM} and Memory Networks \cite{Weston2015MemoryN}, \cite{Sukhbaatar2015EndToEndMN,Miller2016KeyValueMN} are the examples of two such model architectures which incorporates external memory in their learning process. In the context of FSL, memory as an external component can relieve the burden of training in low data regime and can allow for faster generalization. Next, we discuss such model architectures which integrates memory into their design and use meta-learning for learning from few-examples. 

\textbf{Memory Augmented Neural Networks} \\
Memory Augmented Neural Networks (MANN) \cite{Santoro2016OneshotLW} use a modified NTM to rapidly assimilate new data into memory and leverage this data to make accurate predictions after only a few samples.

\begin{itemize}
    \item \textbf{Neural Turing Machine} \\
    A Neural Turing Machine (NTM) couples a controller neural network with external memory storage (Figure \ref{fig:NTM}). Like most neural networks, the controller interacts with the external world via input and output vectors. Unlike standard networks, the controller also learns to read and write memory rows by soft attention, while the memory serves as a knowledge repository. The attention weights are generated by its addressing mechanism.
    \begin{figure*}[ht]
        \centering
        \includegraphics[keepaspectratio,width=0.5\textwidth]{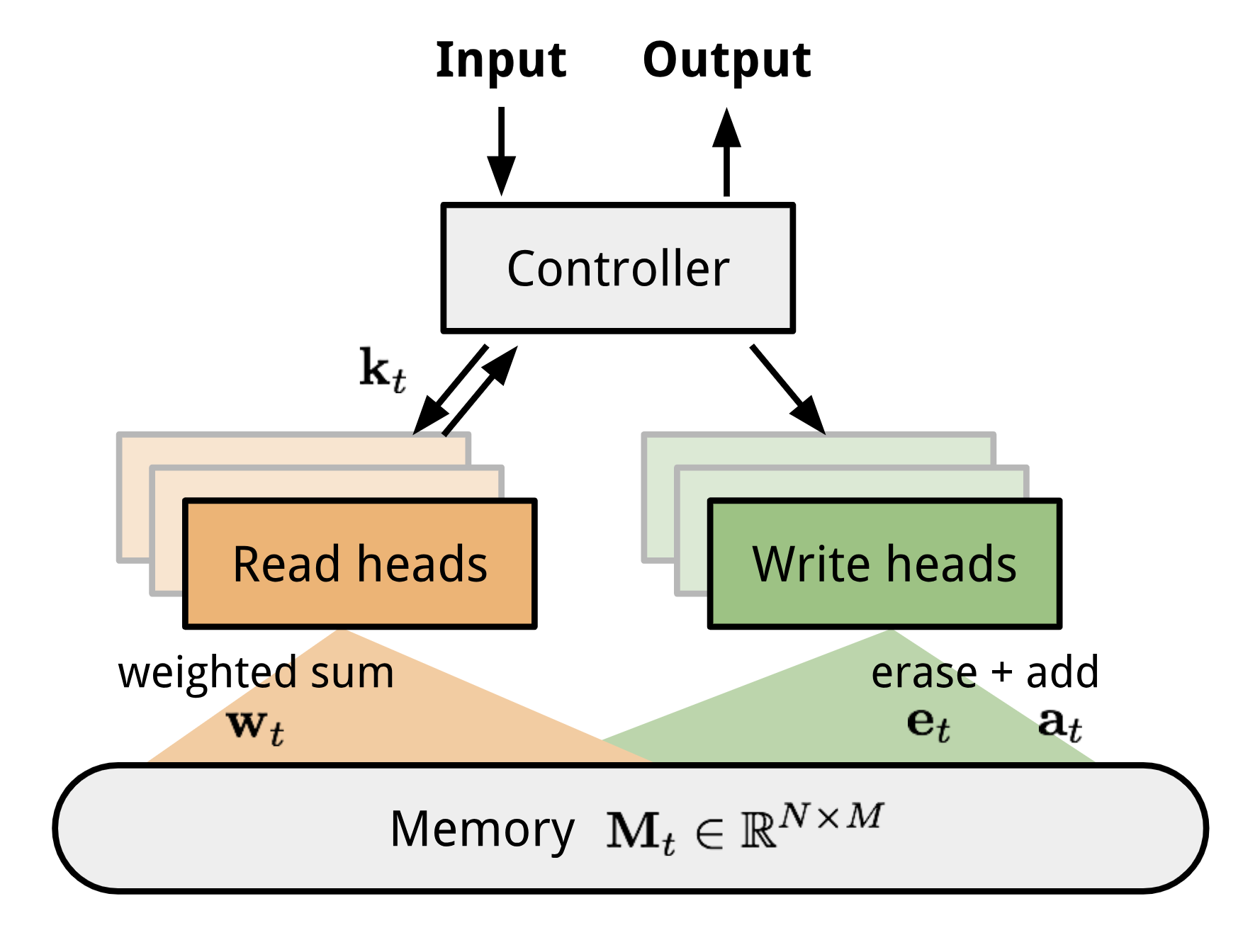}
        \caption{The architecture of NTM. The memory at time t, $M_t$ is a matrix of size $N \times M$, containing N vector rows and each has M dimensions (Source:  \cite{lilianweng}).}
        \label{fig:NTM}
    \end{figure*}
    \item \textbf{Addressing Mechanism in MANN} \\
    The controllers in MANN are either LSTMS or feed-forward networks. Given some input $x_t$ at time $t$, the controller produces a key feature vector $k_t$, which is then either stored in a row of a memory matrix $M_t$, or used to retrieve a particular memory, i, from a row; i.e. $M_t(i)$. A memory, $r_t$, is retrieved using the weighting vector $w_t^r(i)$ as:
    $$ r_t \leftarrow \sum_i w_t^r(i) M_t(i) \text{ where } w_t^r(i) = \text{softmax}(\frac{k_t \cdot M_t(i)}{||k_t|| \cdot ||M_t(i)||}).$$ 
    Writing to memory in MANN model involves the use of Least Recently Used Access (LRUA) module. The LRUA module is a pure content-based memory writer that writes memories to either the least used memory location or the most recently used memory location.
    \item \textbf{Meta-Learning Setup for MANN} \\
    Memory encoding and retrieval in a NTM external memory is rapid, with vector representations being replaced into or taken out of memory potentially every time-step. This ability makes the NTM a perfect candidate for meta-learning and low-shot prediction, as it is capable of both long-term storage via slow update of weights, and short term storage via its external memory module. Training in MANN follows the same episodic paradigm discussed earlier, except the truth label $y_t$ is presented with one step offset i.e $\{(x_t, y_{t-1}),(x_{t+1}, y_{t}), ...\}$. The network is tasked to output the appropriate label for $x_t (i.e., y_t)$ at the given timestep. This prevents the network from slowly learning sample-class bindings in its weights. Instead, it must learn to hold data samples in memory until the appropriate labels are presented at the next time-step, after which sample-class information can be bound and stored for later use (Figure \ref{fig:MANN}). 
    \begin{figure*}[ht]
        \centering
        \includegraphics[keepaspectratio,width=\textwidth]{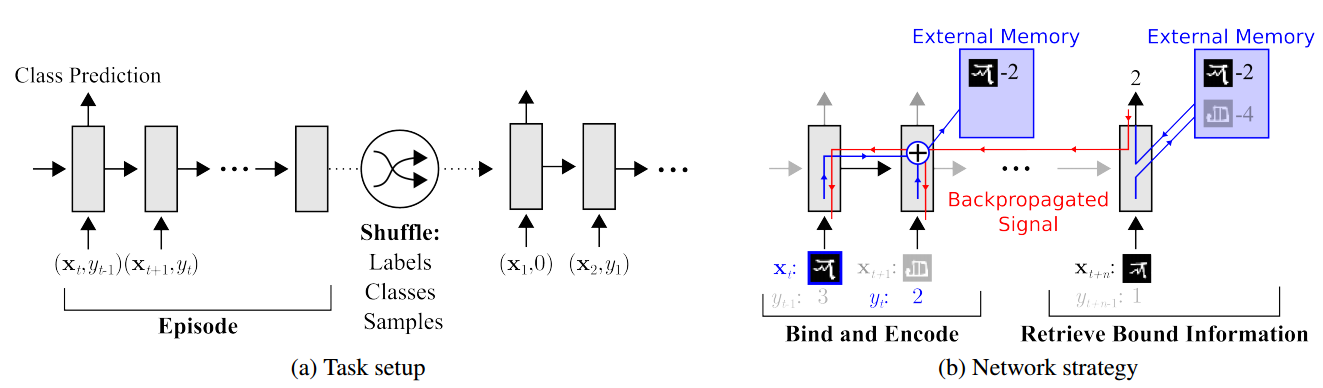}
        \caption{Task Structure. (a) Images $x_t$ are presented with time-offset labels, (b) External memory stored the bounded sample representation-class label information (Source: \cite{Santoro2016OneshotLW}).}
        \label{fig:MANN}
    \end{figure*}
\end{itemize}

\noindent \textbf{Memory Matching Networks} \\
Memory Matching Networks (MM-Net) \cite{Cai2018MemoryMN} integrates the memory module from Key-Value Memory Networks \cite{Miller2016KeyValueMN} into Matching Networks \cite{Matching2016}. It extends the idea of metric-based meta-learning with a memory module to encode and generalize the whole support set into memory slots. Given the support set $S$, the memory module encodes the sequence of $N$ support images into $M$ memory slots with the \textit{write controller}. For each support image $x$ and its embedded representation $z$ in memory key space, the \textit{read controller} measures the dot product similarity between the input support image and the memory slots to retrieve a conditioned representation $g(x|M)$. Meanwhile, a contextual learner is devised to predict the parameters of CNNs for embedding unlabeled image in the query set.

\begin{figure*}[ht]
    \centering
    \includegraphics[keepaspectratio,width=0.9\textwidth]{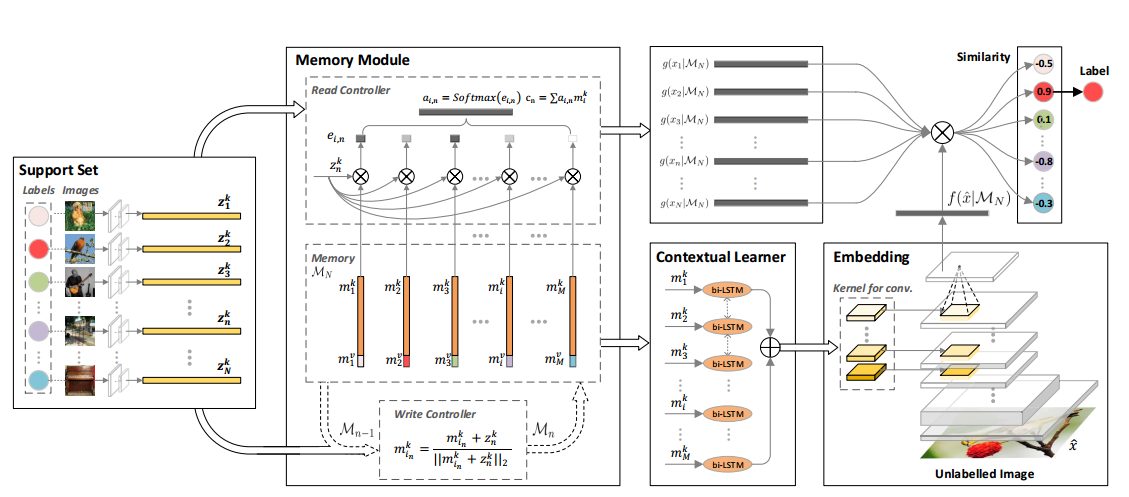}
    \caption{Architecture of Memory Matching Networks (Source: \cite{Cai2018MemoryMN})}
    \label{fig:MMNet}
\end{figure*}

\subparagraph{Rapid Adaptation} \mbox{} \\
 The following model-based approaches uses techniques like "fast-weights" to rapidly adapt the parameters of a model for a given task. Normally weights in the neural networks are updated by stochastic gradient descent in an objective function and this process is known to be slow. One faster way to learn is to utilize one neural network to predict the parameters of another neural network and the generated weights are called fast weights. In comparison, the ordinary SGD-based weights are named slow weights.

\textbf{Meta Networks} \\
Meta Networks (MetaNet) \cite{Munkhdalai2017MetaN} is a meta-learning model with architecture and training process designed for \textit{rapid} generalization across tasks. It consists of two main learning components, a base learner and a meta-learner, and is equipped with an external memory. The rapid generalization of MetaNet relies on fast-weights.  The external memory is used to store these fast weights and the input representations. 

In MetaNet, loss gradients are used as meta information to populate models that learn fast weights. Slow and fast weights are combined to make predictions in neural networks.

\begin{figure*}[ht]
    \centering
    \includegraphics[keepaspectratio,height=0.4\textwidth]{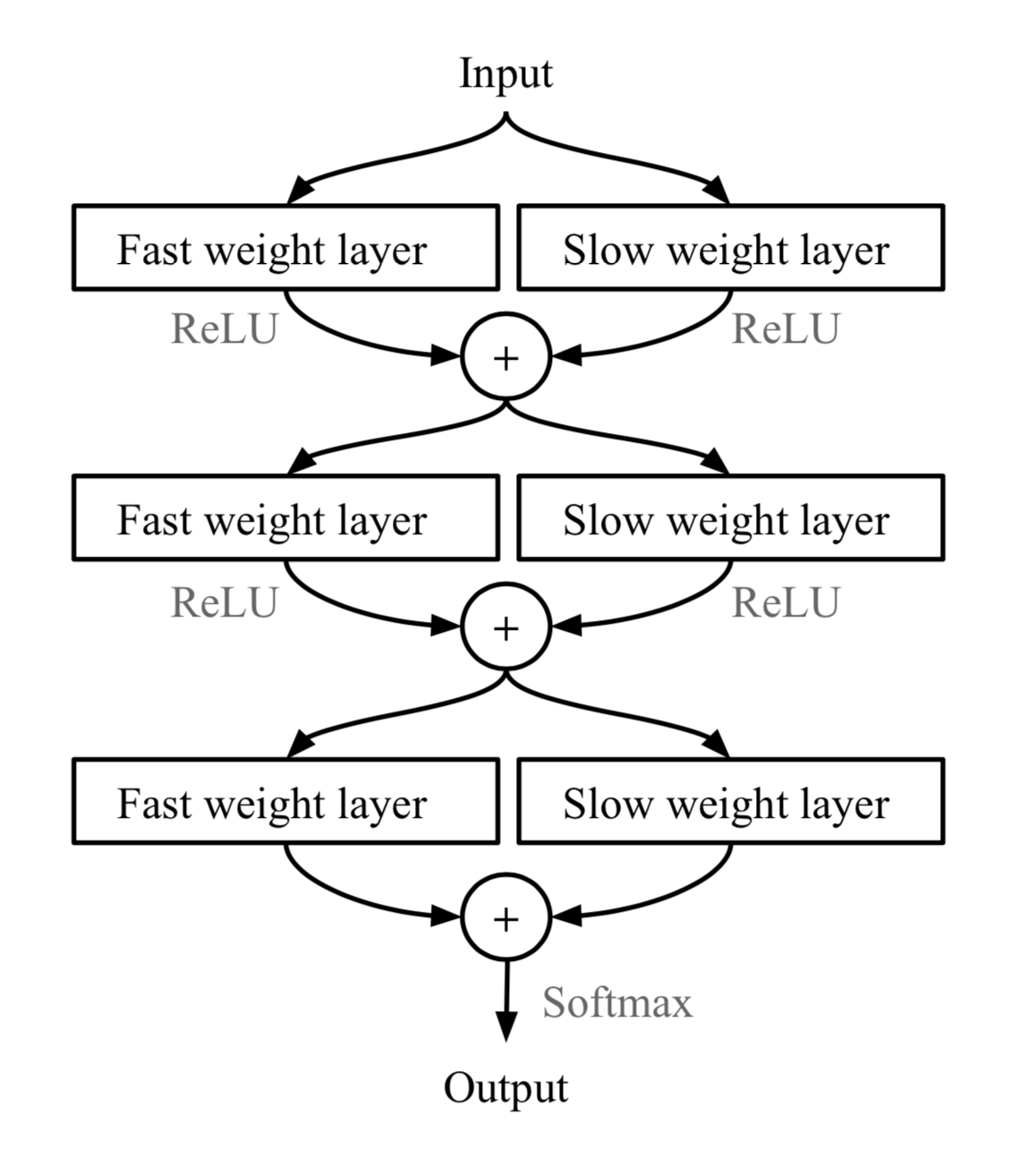}
    \caption{Combining slow and fast weights in a MLP. $\bigoplus$ is element-wise sum (Source: \cite{Munkhdalai2017MetaN}).}
    \label{fig:MetaNet}
\end{figure*}

\textbf{Conditionally Shifted Neurons} \\
Conditionally Shifted Neurons (CSNs) \cite{Munkhdalai2017RapidAW} modify their activation values with task-specific shifts retrieved from a memory module, which is populated rapidly based on a limited task experience.  Building upon Meta-Networks, CSNs also have a base learner, a meta-learner and a memory module. The learning happens in the following manner:  

\begin{itemize}
    \item A base learner works on individual tasks. Using its current weights it makes predictions on the examples in the support set (description phase).
    
    \item The loss incurred from each prediction from support set is stored in the form of conditioning information $I$ in the key-value memory where keys are the embeddings of inputs and the values are the conditioning information. 

    \item To classify a query (prediction phase), its embeddings are compared with the embeddings of the keys in the memory using cosine similarity and the similarity score is then weighted using a softmax.  The conditioning information for each key is weighted by its similarity score and summed together to obtain joint conditioning information.
    
    \item The base network is updated with this joint conditioning information and the prediction for the query is made with these updated weights.

    \item The loss obtained on the query is then used to update the key embeddings network $f$, the value network $g$ and the prediction network (the initial base network).
\end{itemize}

\begin{figure*}[ht]
    \centering
    \includegraphics[keepaspectratio,width=0.5\textwidth]{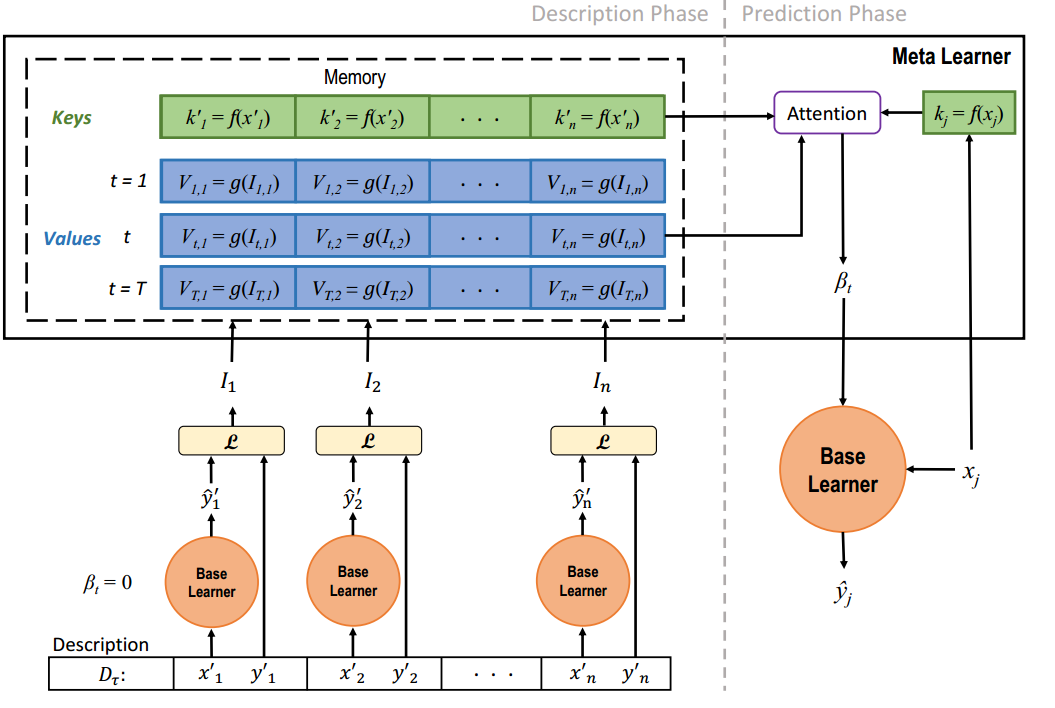}
    \caption{Rapid Adaptation with Conditionally Shifted Neurons. In the description phase, the meta learner populates
working memory with keys and values, based on the base learner’s performance on the task description; in the prediction phase, the meta learner retrieves task-specific shifts from memory through key-based attention and feeds them to the base learner to adapt it to the task  (Source: \cite{Munkhdalai2017RapidAW}).}
    \label{fig:CSN}
\end{figure*}

\begin{figure*}[ht]
    \centering
    \includegraphics[keepaspectratio,height=0.4\textwidth]{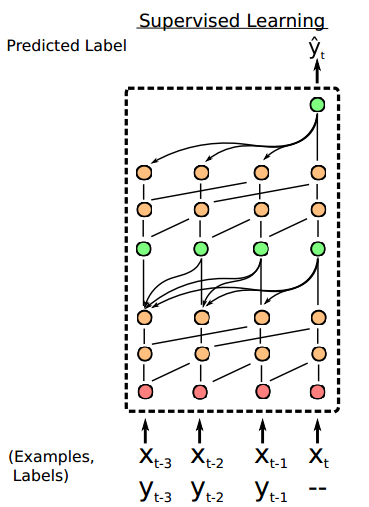}
    \caption{Overview of SNAIL; in this example, two blocks of
temporal convolution layers (orange) are interleaved with two causal attention layers (green) (Source: \cite{Mishra2017ASN}).}
    \label{fig:SNAIL}
\end{figure*}

\subparagraph{Miscellaneous Models: SNAIL} \mbox{} \\
Each episode in SNAIL \cite{Mishra2017ASN} receives as input a sequence of example-label pairs $(x_1,y_1), ... ,(x_{t-1}, y_{t-1})$ for timesteps 1, ..., $t-1$, followed by an unlabeled example $(x_t, \_ )$ (Figure \ref{fig:SNAIL}). It is then tasked to output prediction for $x_t$ based on the previous labeled examples it has seen.
Formalizing meta-learning as a sequence-to-sequence problem, it argues that meta-learner should be able to internalize and refer to past experience. It proposes use of embedding networks with interleaved temporal convolutions and causal attention layers; the former to aggregate information from past experience and the latter to pinpoint specific pieces of information. The temporal convolution layers in SNAIL provides high-bandwidth access over its past experience without constraints on the amount of experience it can effectively use. SNAIL architectures are easier to train than traditional RNNs such as LSTM or GRUs and can be efficiently implemented so that an entire sequence can be processed in a single forward pass. 

\subsubsection{Hybrid Approaches} \label{sec:hybrid}
This section discusses the variations of the few-shot learning problem and the hybrid meta-learning based approach towards them. Table \ref{table:hybrid} list the hybrid approaches and summarizes the key idea behind them. 

\begin{table}[!ht]
    \centering
    \begin{tabular}{lp{10cm}}
    \hline
        \textbf{Approach} & \textbf{Key Idea} \\ \hline
    Cross-Modal FSL & Leverage semantic data from a \textbf{different modality}. \\ 
    Semi-Supervised FSL & Using few training examples, label the given \textbf{unlabeled} examples and improve the few-shot classifier. \\ 
    Generalized FSL & Classify query into \textbf{either} one of the meta-training classes  or a class from the support set. \\ 
    Generative FSL  & Learning to \textbf{generate} more samples from the given few. \\ 
    Cross Domain FSL & Training in \textbf{one domain} and testing \textbf{in another}. \\  
    Transductive FSL & \textbf{Jointly predict} all the query examples. \\ 
    Unsupervised FSL & Support examples are \textbf{unlabeled}. \\ 
    Zero-Shot Learning & \textbf{No support examples} present. \\ 
    \hline
    \end{tabular}
    \caption{Hybrid Approaches}
    \label{table:hybrid}
\end{table}

\textbf{Cross-Modal Few-Shot Learning} \\ 
The recent progress in few-shot image classification has primarily been made in the context of unimodal learning. In order to alleviate the problem of limited data in image domain, some approaches \cite{Wang2017MultiattentionNF,Xing2019AdaptiveCF} employ data from different modality (eg., text). This is referred to as Cross-Modal Few-Shot Learning. For example, Xing et al. \cite{Xing2019AdaptiveCF} propose an Adaptive Modality Mixture Mechanism (AM3) that combines information from an image and its label's word embedding to develop a better prototype of its class.

\textbf{Semi-Supervised Few-Shot Learning} \\
Semi-Supervised FSL considers the scenario when there is limited \textit{labeled} data but sufficient \textit{unlabeled} data is available during the training. For example in approaches which use meta-learning for few-shot classification, a weakly supervised classifier trained on support set (labeled) learns to label unlabeled data and then use it to improve its performance on the classification task. Using metric-based meta-learning approach, Ren et al. \cite{Ren2018MetaLearningFS} propose three semi-supervised variants of Prototypical Networks \cite{Snell2017PrototypicalNF}, basically using Soft k-Means method to tune clustering centers with unlabeled data. Alternatively, Sun et al. \cite{Sun2019LearningTS} use an optimization-based meta-learning approach for learning to initialize a classification model for semi-supervised FSL.

\textbf{Generalized Few-Shot Learning} \\ 
In general, FSL approaches involve meta-training on base (seen) classes and meta-testing on novel (unseen) classes.  Given a novel task sampled from the meta-testing set, the few-shot classifier will classify a query into one of the classes present in the novel task's support set (task's training set) and will not be able to recognize if the query example is from a base class. Generalized Few-Shot Learning (GFSL) focuses on the \textit{joint} classification of both the base classes and novel classes. In particular, the goal is, for the model trained on the seen categories to be capable of incorporating the limited unseen class instances and make predictions for test/query instances in the both set of classes. Recent work of Gidaris \& Komodakis \cite{Gidaris2018DynamicFV}, Ye et al. \cite{Ye2019LearningCS} and Ren et al. \cite{Ren2019IncrementalFL} attempts to address this problem.

\textbf{Generative Few-Shot Learning} \\ 
One usual way one may think to alleviate the problem of learning from low data samples is to augment them with synthesized samples.  To this end, Wang et al. \cite{Wang2018LowShotLF} propose a meta-learning approach for generating samples from limited data. This generative model is referred to as \textit{hallucinator}, a model that maps real examples to hallucinated examples. The few-shot training set is first fed to the hallucinator, and it produces an expanded training set, which is then used by the learner. We refer to approaches of this kind as Generative Few-Shot Learning. 

\textbf{Cross Domain Few-Shot Learning}  \\
The FSL approaches discussed so far in the context of few-shot classification aimed to recognize novel categories with only few labeled examples in each class. One assumption that was followed was that all few-shot tasks belong to the same distribution or domain. For example, most approaches sampled tasks from \textit{mini}ImageNet during training and as well as testing. While promising results were observed following this assumption, the existing methods often fail to generalize to unseen domains due to large discrepancy of the feature distribution across domains. This problem of few-shot learning under domain shifts is referred to as Cross Domain FSL.  To this end, the early works of Tseng et al. \cite{Tseng2020CrossDomainFC} attempt to address this problem by simulating various feature distributions under different domains in the training stage.

\textbf{Transductive Few-Shot Learning} \\  
Although, meta-learning is an effective strategy for few-shot learning as it aims at generalizing to unseen classification tasks, the fundamental difficulty with learning with scarce data remains for a novel classification task. One way to achieve larger improvements with limited amount of training data is to consider relationships between instances in the test set and thus predicting them as whole, which is referred to as transduction or transductive inference \cite{Vapnik2006TransductiveIA}. Therefore, transductive few-shot learning techniques \cite{Hou2019CrossAN,Liu2019LearningTP,Antoniou2019LearningTL} utilize the information present in unlabeled examples in the query set as whole to make prediction about individual queries. For example Liu et al. \cite{Liu2019LearningTP} proposed Transductive Propagation Networks, where the examples in the support and query sets are modeled as nodes of a graph. The labels of the support set nodes are known and the task is to predict the labels of the query set nodes which is achieved using their label propagation algorithm.

\textbf{Unsupervised Few-Shot Learning} \\
In supervised few-shot learning, the labels of the examples in the support set are available during the training and the label of the examples in the query set have to be estimated. In contrast, in unsupervised few-shot learning, the examples in the support set are also unlabeled. Huang et al. \cite{Huang2019CentroidNF} proposed a strategy for unsupervised few-shot classification task which involves first performing clustering on the examples in the support set and then assigning the query to one of the clusters.

\textbf{Zero-Shot Learning} \\
Zero-Shot Learning (ZSL) \cite{ZSL} attempts to solve a task without the presence of any training examples for that task. For an image classification task, ZSL methods rely mostly on visual-auxiliary modality alignment. Often times, the auxiliary data is image's label i.e., samples for the same class from two modalities are mapped together so that two modalities obtain the same semantic structure. Because ZSL does not have access to any visual information when learning new concepts, ZSL models have no choice but to align the two modalities. This way, during test the image query can be directly compared to auxiliary information for performing classification \cite{Zhang2017LearningAD}.

\subsection{Non-Meta-Learning based Few-Shot Learning} \label{sec:non-meta-fsl}
In this section, we discuss strategies other than meta-learning that can aid learning in limited data regime.

\subsubsection{Transfer Learning}
Transfer Learning \cite{torrey2010transfer} is the improvement of learning in a new task through the transfer of knowledge from a related task that has already been learned. In few-shot learning scenario where the data is too limited to train a deep network from scratch,  transferring knowledge from another network can be a viable option. For a classification task, this knowledge transfer is achieved by pretraining a deep network on large amounts of training data on base classes (seen) and then fine-tuning it on a new few-shot classes (unseen).  However, naive fine-tuning using just few example can lead to overfitting and thus poor generalization performance on the few-shot task. Therefore, in this section, we discuss approaches which attempts to address this problem.

\textbf{Distance Metric Classification Using Embeddings from a Pretrained Network} \\
In section \ref{sec:metric}, we discussed approaches that used meta-learning to extract feature embeddings and perform classification using a nearest neighbour classifier with a distance metric. SimpleShot \cite{Wang2019SimpleShotRN} instead uses a pretrained deep network to get feature embeddings for the input and query images, perform centering \& L2 normalization on the obtained features and use euclidean distance as the distance measure for nearest neighbour classification. This simple approach has shown considerable improvement in accuracy in comparison to meta-learning approaches. Similarly chen et al. \cite{Chen2020ANM}, shows comparable results from using cosine metric for nearest neighbour classification on embeddings obtained from a network trained on base classes. 

\textbf{Training a New Classifier Using Embeddings from a Pretrained Network} \\
Training a classifier from scratch isn't possible when the number of training samples are limited, due to the poor resulting representations. However one could still obtain the representations from a pre-trained network and then train a new classifier using them. Tian et al. \cite{tian2020rethinking}, demonstrate this exactly. Along with using representations from a pre-trained network, they also L2 normalize them before training a new classifier for each few-shot task. They also show that this approach outperforms the simple nearest neighbour classification using pretrained embeddings.

\textbf{Transductive Inference Using Embeddings from a Pretrained Network} \\
Certain approaches attempt to exploit the structure of information present in the query set and collectively classify the examples in the query set. This is know as transductive inference. For example, in section \ref{sec:hybrid}, we mentioned Transductive Propagation Networks \cite{Liu2019LearningTP} which utilized meta-learning to  transductively assign labels to the examples in the query set. Alternatively, instead of using meta-learning, Dhillon et al. \cite{Dhillon2019ABF} choose to transductively fine-tune a pretrained network on a given few-shot task. That means along with the support examples (labeled), query samples are also utilized in the fine-tuning process. The proposed transductive fine-tuning phase solves for:

\begin{equation} \label{eq:transFT}
   \theta^* =  \text{arg} \min\limits_{\theta} \frac{1}{|S|} \sum_{(x,y) \in S} -\log p_{\theta} (y|x)  + \frac{1}{|Q|} \sum_{(x,y) \in Q} \mathbb{H}(p_{\theta}(. | x)). 
\end{equation}

The first term in the equation is the data fitting term using labeled support samples whereas the second term, the regularizer, uses the unlabeled query samples to minimize the entropy of predictions. 

Similarly, Ziko et al. \cite{Ziko2020LaplacianRF}, propose a transductive laplacian-regularized inference for few-shot tasks. Using the feature embeddings learned from the base classes (pretrained), they minimize a quadratic binary-assignment function containing two terms:

\begin{equation} \label{eq:objective}
   \mathcal{E}(\mathbf{Y}) = \mathcal{N}(\mathbf{Y}) + \frac{\lambda}{2} \mathcal{L}(\mathbf{Y}),  
\end{equation}

where 

$$\mathcal{N}(\mathbf{Y}) = \sum_{q=1}^N \sum_{c=1}^C y_{q,c}  d(x_q - \mathbf{m}_c)$$

and 

$$\mathcal{L}(\mathbf{Y}) = \frac{1}{2} \sum_{q,p} w(x_q, x_p) || \mathbf{y}_q - \mathbf{y}_p||^2$$

\begin{itemize}
    \item $\mathcal{N}(\mathbf{Y})$, a unary term, is minimized globally when each query point is assigned to the class of the nearest prototype $\mathbf{m}_c$ (obtained from the support set) using a distance metric $d(x_q, m_c)$.
    \item $\mathcal{L}(\mathbf{Y})$, a pairwise Laplacian term, encourages nearby points $(x_p, x_q)$ in the label space to the same latent label assignment ($w$ is any similarity metric).
\end{itemize}

These transfer learning based methods have often come to exhibit better or equivalent performance on FSL tasks when compared to the complex meta-learning methods discussed earlier. 

\subsubsection{Miscellaneous: Autoencoders}
Mocanu \cite{Mocanu2018OneShotLU} proposes a one-shot learning method, dubbed MoVAE (Mixture of Variational Autoencoders), to perform classification. A Variational Autoencoder (VAE) \cite{kingma2013auto} provides a probabilistic manner for describing an observation in latent space. Thus, rather than building an encoder which outputs a single value to describe each latent state attribute, a VAE encoder to describe a probability distribution for each latent attribute. Given C classes, C VAEs are trained, one for each class. Then the reconstruction loss from the VAEs are measured to perform classification on the unlabeled data sample. A downside of this approach is that requires constructing and training new Autoencoders for each new task even at test. In contrast, most meta-learning methods work out-of-the-box at test time.

\pagebreak

\section{Progress in Few-Shot Learning} \label{sec:progress}
Early few-shot research focused on computer vision applications and mainly image classification \cite{Matching2016,Snell2017PrototypicalNF,Oreshkin2018TADAMTD,MAML2017}. This is because visual information is easy to acquire and has been extensively examined in machine learning. Other computer vision problems such as Object Detection \cite{Chen2018LSTDAL,Kang2018FewShotOD,Fan2019FewShotOD, Schwartz2018RepMetRM} and Segmentation \cite{Michaelis2018OneShotIS} have also received recent attention from the few-shot learning community. Apart from computer vision applications, FSL has been used for fault diagnosis \cite{Zhang2019LimitedDR}, text classification \cite{yu2018diverse,deng2019low}, image colorization \cite{yoo2019coloring} and cold-start item recommendation \cite{Vartak2017AMP,Du2019SequentialSM}. In graph modeling, FSL has been used for node classification \cite{zhou2019meta}, edge labelling \cite{kim2019edge} and relation classification \cite{Han2018FewRelAL}. In audio, its used for few-shot speaker recognition \cite{anand2019few,parnami2020few} and sound recognition \cite{chou2019learning}. Finally, imitation learning \cite{duan2017one} in  robotics and control \cite{MAML2017} in reinforcement learning \cite{sutton2018reinforcement}. 

\begin{figure*}[ht]
    \centering
    \includegraphics[keepaspectratio,width=0.8\textwidth]{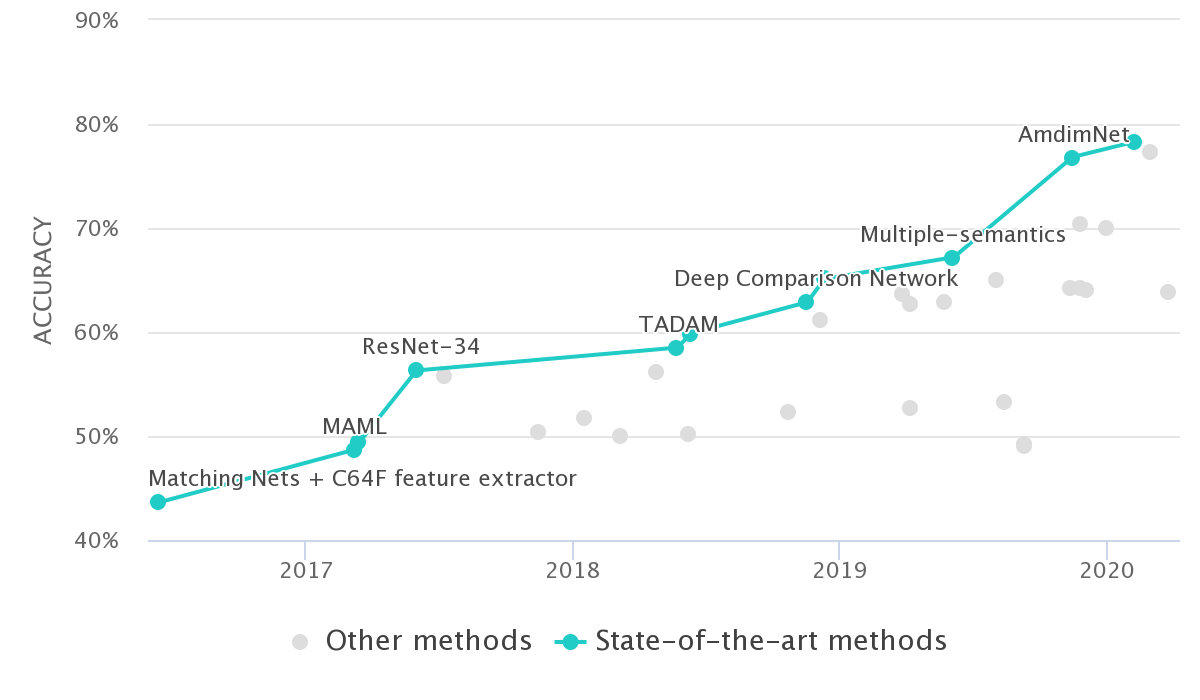}
    \caption{Progress in FSL\protect\footnotemark}
    \label{fig:progress}
\end{figure*}

Since its emergence in 2016, the field of few-shot learning has shown promising improvement in learning from limited data. Figure \ref{fig:progress} shows the trend of improvement in accuracy for 5-way 1-shot classification task on \textit{mini}ImageNet \cite{Matching2016}. Beginning with Matching Networks \cite{Matching2016} at 43\% accuracy, various optimization, metric, model-based and hybrid approaches were proposed in past four years, pushing the accuracy to 80\% (as of Jan 2020). Although model-based approaches have seen less progress, there is no clear consensus if any one particular approach is the best way forward. Table \ref{table:accuracy} lists the accuracy of methods discussed in this survey (sorted by 1-shot accuracy and color coded by type). Metric-based, Optimization-based, Hybrid Meta-Learning and even Non-Meta-Learning approaches, all seem to be leading the race. 

\footnotetext{https://paperswithcode.com/sota/few-shot-image-classification-on-mini-2}

\begin{table}[!ht]
    \centering
    \begin{tabular}{llll}
    \hline
        \textbf{Model} & \textbf{1-shot} & \textbf{5-shot} & \textbf{Type} \\ \hline
\rowcolor{blue!10} Matching Networks \cite{Matching2016} & 43.56  & 55.31  & Metric            \\
\rowcolor{green!10} MAML \cite{MAML2017}              & 48.7   & 63.15  & Optimization      \\
\rowcolor{blue!10} ProtoNet \cite{Snell2017PrototypicalNF}          & 49.42  & 68.2   & Metric            \\
\rowcolor{blue!10} Relation Net \cite{Sung2018LearningTC}      & 50.44  & 65.32  & Metric            \\
\rowcolor{blue!10} ProtoNet + Margin \cite{Zhang2019OneSL} & 51.62  & 70.24  & Metric            \\
\rowcolor{green!10} CAVIA \cite{Zintgraf2019FastCA}             & 51.82  & 65.85  & Optimization      \\
\rowcolor{red!10}  SNAIL \cite{Mishra2017ASN}             & 55.71  & 68.88  & Model            \\
\rowcolor{yellow!10} TPN \cite{Liu2019LearningTP}               & 55.51  & 69.86  & Hybrid            \\
\rowcolor{yellow!10} Dynamic FSL \cite{Gidaris2018DynamicFV}       & 56.2   & 73     & Hybrid            \\
\rowcolor{red!10} CSN \cite{Munkhdalai2017RapidAW}               & 56.88  & 71.94  & Model             \\
\rowcolor{blue!10} SAML \cite{Hao2019CollectAS}              & 57.69  & 73.03  & Metric            \\
\rowcolor{blue!10} TADAM \cite{Oreshkin2018TADAMTD}             & 58.5   & 76.7   & Metric            \\
\rowcolor{green!10} MTL \cite{Sun2018MetaTransferLF}               & 61.2   & 75.5   & Optimization      \\
\rowcolor{blue!10} TapNet \cite{Yoon2019TapNetNN}            & 61.65  & 76.36  & Metric            \\
\rowcolor{green!10} LEO \cite{Rusu2018MetaLearningWL}               & 61.76  & 77.59  & Optimization      \\
\rowcolor{blue!10} CTM \cite{Li2019FindingTF}               & 62.05  & 78.63  & Metric            \\
\rowcolor{green!10} SCA \cite{Antoniou2019LearningTL}               & 62.86  & 77.64  & Optimization      \\
\rowcolor{blue!10} CAN \cite{Hou2019CrossAN}               & 63.85  & 79.44  & Metric            \\
\rowcolor{gray!10} SimpleShot \cite{Wang2019SimpleShotRN}        & 64.29  & \textbf{81.5}   & Non-Meta-Learning \\
\rowcolor{yellow!10} AM3 \cite{Xing2019AdaptiveCF}               & 65.3   & 78.1   & Hybrid            \\
\rowcolor{yellow!10} LST \cite{Sun2019LearningTS}               & 70.1   & 78.7   & Hybrid            \\
    \hline
    \end{tabular}
    \caption{ Accuracy of 5-way classification task on \textit{mini}ImageNet}
    \label{table:accuracy}
\end{table}

\section{Challenges and Open Problems} \label{sec:discussion}

Recent years witnessed a great deal of interest from the machine learning community in solving the problem of learning from few examples. Consequently, many approaches were proposed, and each of them involved some kind of knowledge transfer, either though meta-learning or transfer learning. The success of these approaches relied on certain assumptions, which might become challenging to uphold in real-world settings. Therefore, in this section we discuss the challenges involved in the nitty-gritty of FSL approaches.

\paragraph*{Training the Same Way as Testing} 
Most meta-learning methods employ M-way K-shot episodic training paradigm (Algorithm \ref{alg:episodic}), i.e, we train the few-shot learning model to distinguish between  M classes each with exactly K training examples. This makes the trained few-shot classifier very rigid to be deployed in real world scenarios as one cannot expect to know beforehand the number of classes (M) and the number of training examples (K) available for an unseen task. Moreover, the model will suffer performance degradation if the number of examples present are less than K \cite{Cao2020ATA}. 

\textbf{Learning Constrained to a Single Distribution of Tasks} \\
In few-shot learning experiments, training and testing tasks are sampled from the same distribution $p(\mathcal{T})$, which constraints the learning to a single domain. For example, a FSL model trained for character classification tasks on Omniglot \cite{Lake1332} may not work well with digit classification on MNIST \cite{lecun-mnisthandwrittendigit-2010}. Similarly a general few-shot image classifier built on miniImagenNet \cite{Matching2016} may not work well for fine-grained classification of Birds \cite{WelinderEtal2010} or Cars \cite{CARS}. This problem is also referred to as Cross Domain FSL.

\textbf{Performing Joint Classification from Seen and Unseen Classes} \\   
The classes used in the training phase are not retained by the few-shot classifier, i.e. once the training finishes, the model can only classify the incoming query into one of the classes present in the support set of the given task.  In real-world usage, it would be desirable to jointly classify the incoming query from training classes (seen) and the new classes (unseen) present in the support set of the task. This challenge is also referred to as Generalized Few-Shot Learning.

\textbf{Few-Shot Learning in Data Domains Other than Images }  \\
Availability of large amount of image datasets have resulted in considerable progress in few-shot image classification. Besides the availability, image datasets can be constructed to have uniformly sized images easily categorized into different classes. Contrary to images, data domains like audio and wireless signals  have limited large-scale datasets that are uniformly curated. Since uniformity of a dataset is essential to meta-learning, it is a challenge to deploy FSL methods from images to domains like signals.

\bibliographystyle{unsrt}  
\bibliography{references}  
\end{document}